\documentclass[final,3p,times,doublecolumn]{elsarticle}
\usepackage[mathlines]{lineno}
\usepackage{amsmath,linnum}
\usepackage{amsmath,amssymb}
\usepackage[tight,footnotesize]{subfigure}
\usepackage{multirow}
\usepackage{float}
\usepackage{algorithm}
\usepackage{algorithmic}
\usepackage{dsfont}

\newproof{pf}{Proof}

\journal{Pattern Recognition}

\journal{manuscript}

\begin{document}

\begin{frontmatter}
\title{Learning Locality-Constrained Collaborative Representation for Robust Face Recognition}
\author [scu]{Xi PENG}\ead{pangsaai@gmail.co}
\author [scu]{Lei ZHANG\corref{cor1}}\ead{leizhang@scu.edu.cn}
\author [scu]{ZHANG Yi}\ead{zhangyi@scu.edu.cn}
\author [nus]{Kok Kiong Tan}\ead{kktan@nus.edu.sg}
\address [scu]{Machine Intelligence Laboratory, College of Computer Science, Sichuan University, Chengdu, 610065, China.}
\address [nus]{Department of Electrical and Computer Engineering, National University of Singapore, Engineering Drive 3, Singapore 117576.}
\cortext[cor1]{Corresponding author}

\begin{abstract}
The model of low-dimensional manifold and sparse representation are two well-known concise models that suggest each data can be described by a few characteristics. Manifold learning is usually investigated for dimension reduction by preserving some expected local geometric structures from the original space to a low-dimensional one. The structures are generally determined by using pairwise distance, e.g., Euclidean distance. Alternatively, sparse representation denotes a data point as a linear combination of the points from the same subspace. In practical applications, however, the nearby points in terms of pairwise distance may not belong to the same subspace, and vice versa. Consequently, it is interesting and important to explore how to get a better representation by integrating these two models together. To this end, this paper proposes a novel coding algorithm, called Locality-Constrained Collaborative Representation (LCCR), which improves the robustness and discrimination of data representation by introducing a kind of local consistency. The locality term derives from a biologic observation that the similar inputs have similar code. The objective function of LCCR has an analytical solution, and it does not involve local minima. The empirical studies based on four public facial databases, ORL, AR, Extended Yale B, and Multiple PIE, show that LCCR is promising in recognizing human faces from frontal views with varying expression and illumination, as well as various corruptions and occlusions.
\end{abstract}

\end{frontmatter}

\linenumbers
\section{Introduction}
\label{sec:1}
Sparse representation has become a powerful method to address problems in pattern recognition and computer version, which assumes that each data point $\mathbf{x} \in \mathds{R}^m$ can be encoded as a linear combination of other points. In mathematically, $\mathbf{x} = \mathbf{D}\mathbf{a}$, where $\mathbf{D}$ is a dictionary whose columns consist of some data points, and $\mathbf{a}$ is the representation of $\mathbf{x}$ over $\mathbf{D}$.
If most entries of $\mathbf{a}$ are zeros, then $\mathbf{a}$ is called a sparse representation. Generally, it can be achieved by solving
\begin{equation*}
\label{equ:1}
 (P_0):\hspace{8mm} \mathrm{min}\|\mathbf{a}\|_0  \hspace{3mm}  \mathrm{s.t.} \hspace{3mm} \mathbf{x} = \mathbf{Da},
\end{equation*}
where $\|\cdot\|_0$ denotes $\ell^0$-norm by counting the number of nonzero entries in a vector. $P_0$ is difficult to solve since it is a NP-hard problem. Recently, compressive sensing theory~\cite{Candes2005-Decoding, Donoho2006-large} have found that the solution of $P_0$ is equivalent to that of $\ell^1$-minimization problem ($P_{1,1}$) when $\mathbf{a}$ is highly sparse.
\begin{equation*}
%\label{equ:2}
 (P_{1,1}):\hspace{8mm} \mathrm{min}\|\mathbf{a}\|_1  \hspace{3mm}  \mathrm{s.t.} \hspace{3mm} \mathbf{x} = \mathbf{Da},
\end{equation*}
where $\ell^1$-norm $\|\cdot\|_1$ sums the absolute value of all entries in a vector. $P_{1,1}$ is convex and can be solved
by a large amount of convex optimization methods, such as basis
pursuit (BP)~\cite{Chen2001-Atomic}, least angle regression
(LARS)~\cite{Efron2004-Least}. In \cite{Yang2010-l1-minimization}, Yang et al. make a
comprehensive survey for some popular optimizers.

Benefiting from the emergence of compressed sensing theory, sparse
coding has been widely used for various tasks, e.g., subspace
learning~\cite{Wong2012Discover,Peng2012}, spectral
clustering~\cite{Cheng2010-Learning,Elhamifar2012-Sparse} and matrix factorization~\cite{Wang2011-Image}. In these works, Wright et al.~\cite{Wright2009-Robust} reported a remarkable method that passes
sparse representation through a nearest feature subspace classifier, named sparse representation based classification (SRC). SRC has achieved attractive performance in robust face recognition and has motivated a large amount of works such as~\cite{Zhang2012Joint,Zhang2013-Simultaneous,He2011-Maximum}. The work implies that sparse representation plays a important role in face recognition under the framework of nearest subspace classification~\cite{Li1999-recognition}.

However, is $\ell^1$-norm based sparsity really necessary to improve the performance of face recognition? Several recent works directly or indirectly examined this problem. Yang et al.~\cite{Yang2012-Beyond} discussed the connections and differences between $\ell^1$-optimizer and $\ell^0$-optimizer for SRC. They show that the success of SRC should attributes to the mechanism of $\ell^1$-optimizer which selects the set of support training samples for the given testing sample by minimizing reconstruction error. Consequently, Yang et al. pointed out that the global similarity derived from $\ell^1$-optimizer but sparsity derived from $\ell^0$-optimizer is more critical for pattern recognition. Rigamonti et al.~\cite{Rigamonti2011-sparse} compared the discrimination of two different data models. One is the $\ell^1$-norm based sparse representation, and the other model is produced by passing input into a simple convolution filter. Their result showed that two models achieve a similar recognition rate. Therefore, $\ell^1$-norm based sparsity is
actually not as essential as it seems in the previous claims. Shi et al.~\cite{Shi2011-recognition} provided a more intuitive approach to
investigate this problem by removing the $\ell^1$-regularization term from the objective function of SRC. Their experimental results showed that their method achieves a higher recognition rate than SRC if the original data is available. Zhang et al.~\cite{Zhang2011-Sparse} replaced the $\ell^1$-norm by the $\ell^2$-norm, and their experimental results again support the views that $\ell^1$-norm based sparsity is not
necessary to improve the discrimination of data representation. Moreover, we have noted that Naseem et al.~\cite{Naseem2010-Linear} proposed Linear Regression Classifier (LRC) which has the same objective function with Shi's work. The difference is that Shi et al. aimed to explore the role of sparsity while Naseem et al. focused on developing an effective classifier for face recognition.

As another extensively-studied concise model, manifold learning is usually investigated for dimension reduction by learning and embedding local
consistency of original data into a low-dimensional representation~\cite{He2005-Neighborhood, Belkin2006-Manifold, Yan2007-Graph}. Local
consistency means that nearby data points share the same properties, which is hardly reflected in linear representation.

Recently, some researchers have explored the possibility of integrating the locality (local consistency) with the sparsity together to produce a better data model. Baraniuk et al.~\cite{Richard2006-Random} successfully bridged the connections between sparse coding and manifold learning, and have founded the theory for random projections of smooth manifold; Majumdar et al.~\cite{Majumdar2010-Robust} investigated the effectiveness and robustness of random projection method in classification task. Moreover, Wang et al.~\cite{Wang2010-Locality} proposed a hierarchal images classification method named locality-constrained linear coding (LLC) by introducing dictionary learning into Locally Linear Embedding (LLE)~\cite{Roweis2000}. Chao et al.~\cite{Chao2011-Locality} presented an approach to unify group sparsity and data locality by introducing the term of ridge regression into LLC; Yang et al.~\cite{Yang2012-Relaxed} incorporated the prior knowledge into the coding process by iteratively learning a weight matrix of which
the atoms measure the similarity between two data points.

\begin{figure*}[t]
\subfigure[]{\label{fig:1.a}\centering\includegraphics[width=0.39\textwidth]{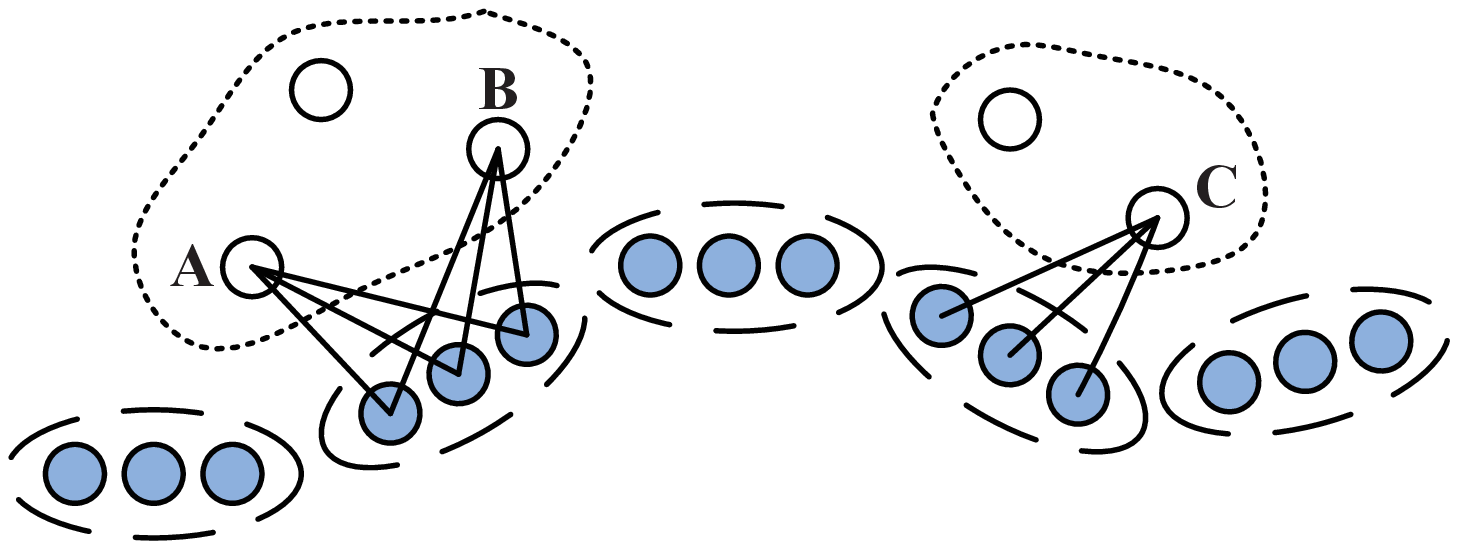}}
\hspace{0.5cm}
\subfigure[]{\label{fig:1.b}\centering\includegraphics[width=0.52\textwidth]{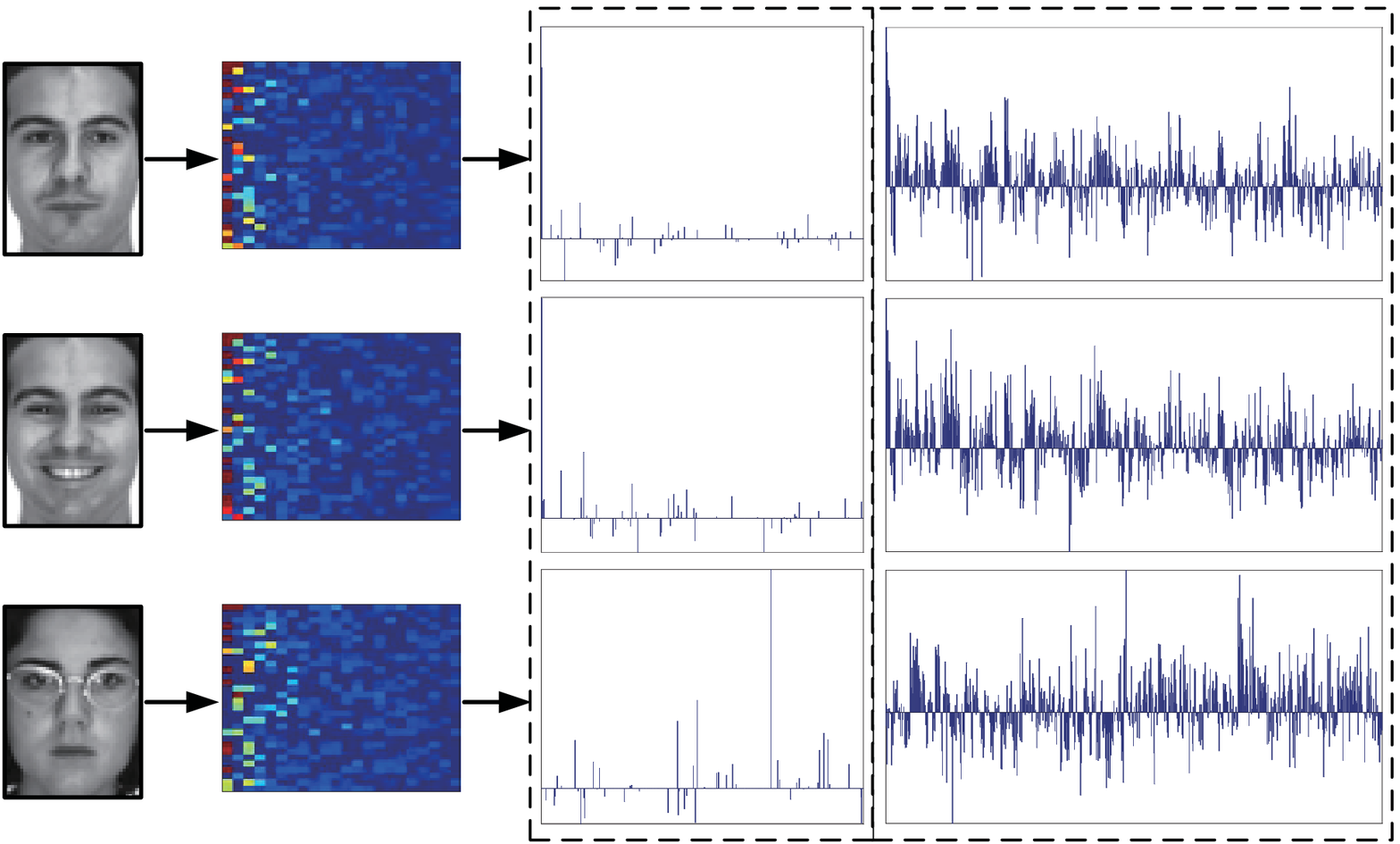}}
\caption{A key observation. (a) Three face images from two different sub-manifolds are linked to their corresponding neighbors, respectively. (b) The first column includes three images which correspond to the points $A$, $B$ and $C$ in \figurename~\ref{fig:1.a}; The second column shows the Eigenface feature matrices for the testing images; The third column includes two parts: the left part is the coefficients of SRC~\cite{Wright2009-Robust}, and the right one is of CRC-RLS~\cite{Zhang2011-Sparse}. From the results, we could see that the representations of nearby points are more similar than that of non-neighboring points, i.e., local consistency could be defined as the similar inputs have similar codes.}
\label{fig:1}
\end{figure*}

In this paper, we proposed and formulated a new kind of local consistency into the linear coding paradigm by enforcing \textbf{the similar inputs (neighbors) produce similar codes}. The idea is motivated by an observation in biological founds~\cite{Ohki2005} which shows that L2/3 of rat visual cortex activates the same collection of neurons in response to leftward and rightward drifting gratings. \figurename~\ref{fig:1} show an example to illustrate the motivation. There are three face images $A$, $B$ and $C$ selected from two different individuals, where $A$ and $B$ came from the same person. This means that $A$ and $B$ lie on the same subspace and could represent with each other. \figurename~\ref{fig:1.b} is a real example corresponding to \figurename~\ref{fig:1.a}. Either from the Eigenface matrices or the coefficients of the two coding schemes, we can see that the similarity between $A$ and $B$ is much higher than the similarity between $C$ and either of them.

Based on the observation, we proposed a representation learning method for robust face recognition, named as Locality-Constrained Collaborative Representation (LCCR), which not aims to obtain a representation that could reconstruct the input with the minimal residual but simultaneously reconstruct the input and its neighborhood such that the codes are as similar as possible. Furthermore, the objective function of LCCR has an analytic solution, does not involve local minima. Extensive experiments show that LCCR outperforms SRC~\cite{Wright2009-Robust}, LRC~\cite{Shi2011-recognition,Naseem2010-Linear}, CRC-RLS~\cite{Zhang2011-Sparse}, CESR~\cite{He2011-Maximum}, LPP~\cite{He2003-Locality}, and linear SVM in the context of robust face recognition.

Except in some specified cases, lower-case bold letters represent column vectors and upper-case bold ones represent matrices, $\mathbf{A}^T$ denotes the transpose of the matrix $\mathbf{A}$, $\mathbf{A}^{-1}$ represents the pseudo-inverse of $\mathbf{A}$, and $\mathbf{I}$ is reserved for identity matrix.

The remainder of paper is organized as follows: Section~\ref{sec:2} introduces three related approaches for face recognition based on data representation, i.e., SRC~\cite{Wright2009-Robust}, LRC~\cite{Shi2011-recognition,Naseem2010-Linear} and CRC-RLS~\cite{Zhang2011-Sparse}. Section~\ref{sec:3} presents our LCCR algorithm. Section~\ref{sec:4} reports the experiments on several facial databases. Finally, Section~\ref{sec:5} contains the conclusion.

\section{Preliminaries}
\label{sec:2}

%\subsection{Notation}
%\label{sec:2.1}

We consider a set of $N$ facial images collected from $L$
subjects. Each training image, which is denoted as a vector
$\mathbf{d}_i \in \mathds{R}^{M}$, corresponds to the $i$th column
of a dictionary $\mathbf{D}\in \mathds{R}^{M\times N}$. Without
generality, we assume that the columns of $\mathbf{D}$ are sorted
according to their labels.
%For a testing image $\mathbf{x}\in \mathds{R}^{M}$, let $\mathbf{a}\in \mathds{R}^{N}$ represents its optimal coefficient on $\mathbf{D}$.

\subsection{Sparse representation based classification}
\label{sec:2.2}

Sparse coding aims at finding the most sparse
solution of $P_{1,1}$. However, in many practical problems, the
constraint $\mathbf{x} = \mathbf{Da}$ cannot hold exactly since
the input $\mathbf{x}$ may include noise. Wright et
al.~\cite{Wright2009-Robust} relaxed the constraint to $\|\mathbf{x} -
\mathbf{Da}\|_2 \leq \varepsilon$, where $\varepsilon >0$ is the
error tolerance, then, $P_{1,1}$ is rewritten as:
\begin{equation*}
\label{equ:3}
(P_{1,2}):\hspace{6mm} \mathrm{min}\|\mathbf{a}\|_1
\hspace{3mm}  \mathrm{s.t.} \hspace{3mm} \|\mathbf{x} -
\mathbf{Da}\|_2 \leq \varepsilon.
\end{equation*}

Using Lagrangian method, $P_{1,2}$ can be transformed to the following
unconstrained optimization problem:
\begin{equation*}
\label{equ:5}
(P_{1,3}):\hspace{6mm}
\mathop{\mathrm{argmin}}_{\mathbf{a}}\|\mathbf{x}-\mathbf{Da}\|_2^2+\lambda
\|\mathbf{a}\|_1,
\end{equation*}
where the scalar $\lambda\geq0$ balances the importance between
the reconstruction error of $\mathbf{x}$ and the sparsity of code
$\mathbf{a}$. Given a testing sample $\mathbf{x}\in \mathds{R}^M$,
its sparse representation $\mathbf{a^*}\in \mathds{R}^N$ can be
computed by solving $P_{1,2}$ or $P_{1,3}$.

After getting the sparse representation of $\mathbf{x}$, one
infers its label by assigning $\mathbf{x}$ to the class that has
the minimum residual:
\begin{equation}
\label{equ:6}
r_i(\mathbf{x})=\|\mathbf{x}-\mathbf{D}\cdot\delta_i(\mathbf{a^*})\|_2,
\end{equation}
\begin{equation}
\label{equ:7}
\mathrm{identity}(\mathbf{x})=\mathop{\mathrm{argmin}}_{i}\{r_i(\mathbf{x})\}.
\end{equation}
where the nonzero entries of $\delta_i(\mathbf{a^*})\in
\mathds{R}^N$ are the entries in $\mathbf{a^*}$ that are
associated with $i$th class, and $\mathrm{identity}(\mathbf{x})$
denotes the label for \textbf{x}.

\subsection{$\ell^2$-minimization based methods}
\label{sec:2.3}

In~\cite{Naseem2010-Linear}, Naseem et al. proposed a Linear Regression Classifier (LRC) which achieved comparable accuracy to SRC in the context of robust face recognition. In another independent work\cite{Shi2011-recognition}, Shi et al. used the same objective function with that of LRC to discuss the role of $\ell^1$-regularization based sparsity. The objective function used in~\cite{Naseem2010-Linear, Shi2011-recognition} is
\begin{equation*}
\label{equ:8}
\mathop{\mathrm{argmin}}_{\mathbf{a}}\|\mathbf{x}-\mathbf{Da}\|_2^2.
\end{equation*}

In~\cite{Shi2011-recognition}, Shi et al. empirically showed that their method (denoted as LRC in this paper for convenience) requires $\mathbf{D}$ to be an over-determined matrix for achieving competitive results, while the dictionary $\mathbf{D}$ of SRC must be under-determined according to compressive sensing theory. Once the optimal code $\mathbf{a^*}$ is calculated for a given input, the classifier (\ref{equ:6}) and (\ref{equ:7}) is used to determine the label for the input $\mathbf{x}$.

As another recent $\ell^2$-norm model, CRC-RLS~\cite{Zhang2011-Sparse} estimates the representation $\mathbf{a^*}$ for the input
$\mathbf{x}$ by relaxing the $\ell^1$-norm to the $\ell^2$-norm in
$P_{1,3}$. They aimed to solve following objective function:
\begin{equation*}
\label{equ:9}
\mathop{\mathrm{argmin}}_{\mathbf{a}}\|\mathbf{x}-\mathbf{Da}\|_2^2
+ \lambda \| \mathbf{a} \|_2^2,
\end{equation*}
where $\lambda>0$ is a balance factor.

LRC and CRC-RLS show that $\ell^2$-norm based data models
can achieve competitive classification accuracy with hundreds of
times speed increase, compared with SRC. Under this background, we aim
to incorporate the local geometric structures into coding process
for achieving better discrimination and robustness.

\section{Locality-Constrained Collaborative Representation}
\label{sec:3}

\begin{figure*}[t]
\centering
\includegraphics[width=0.9\textwidth]{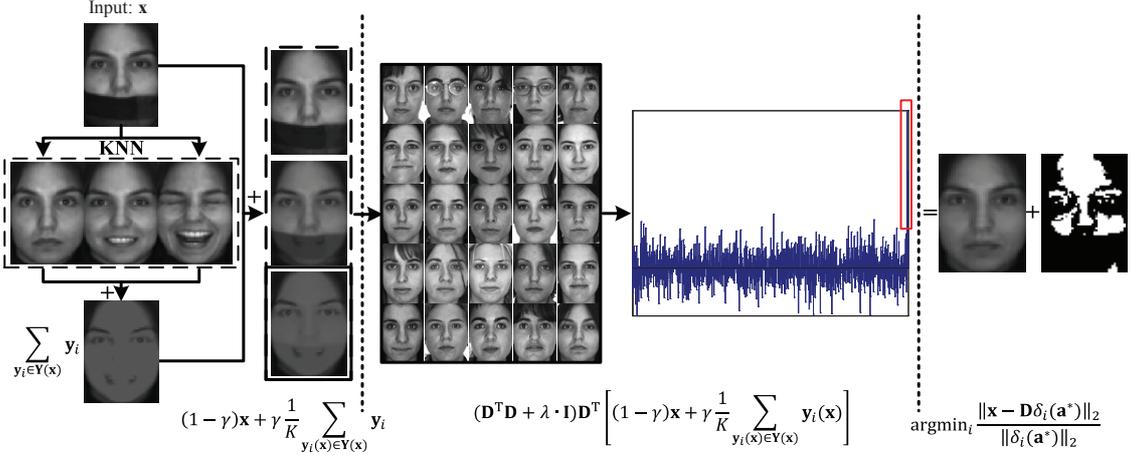}
\caption{Overview of the coding process of LCCR, which consists of three steps separated by dotted lines. First, for a given input $\mathbf{x}$, find its neighborhood $\mathbf{Y(x)}$ from training data. Then, code $\mathbf{x}$ over $\mathbf{D}$ by finding the optimal representation $\mathbf{a}$ (see bar graph) which produces the minimal reconstruction errors for $\mathbf{x}$ and $\mathbf{Y(x)}$ simultaneously. Finally, conduct classification by finding which class produces the minimum residual. In the middle part of the figure, we use a red rectangles to indicate the basis vectors which produce the minimum residual.}
\label{fig:2}
\end{figure*}

It is a big challenge to improve the discrimination and the
robustness of facial representation because a practical face
recognition system requires not only a high recognition rate but
also the robustness against various noise and occlusions.

\subsection{Algorithm Description}

As two of the most promising methods, locality preservation based algorithm and sparse representation have been extensively studied and successfully applied to appearance-based face recognition, respectively. Locality preservation based algorithm aims to find a low-dimensional model by learning and preserving some properties shared by nearby points from the original space to another one. Alternatively, sparse representation, which encodes each testing sample as a linear combination of the training data, depicts a global relationship between testing sample with training ones. In this paper, we aim to propose and formulate a kind of local consistency into coding scheme for modeling facial data. Our objective function is in the form of
\begin{equation}
\label{equ:10}
\mathrm{E}(\mathbf{x},\mathbf{a}) =
\|\mathbf{x}-\mathbf{Da}\|_2^2 + \lambda \| \mathbf{a} \|_p +
\gamma\mathrm{E_L},
\end{equation}
where $p=\{1, 2\}$, $\mathrm{E_L}$ is the locality constraint,
$\lambda\geq0$ and $\gamma\geq0$ dictate the importance of
$\|\cdot\|_p$ and  $\mathrm{E_L}$, respectively. Then the key is
to formulate the shared property of the neighborhood with $\mathrm{E_L}$.

$\mathrm{E_L}$ could be defined as the reconstruction
error of the neighborhood of the testing image, i.e.,
\begin{equation}
\label{equ:11}
\mathrm{E_L}(\mathbf{Y(x)}, \mathbf{C}) =
\frac{1}{K}
\sum_{\substack{\mathbf{y}_i(\mathbf{x})\in \mathbf{Y(x)} \\
\mathbf{c}_i\in
\mathbf{C}}}\|\mathbf{y}_i(\mathbf{x})-\mathbf{Dc}_i\|_2^2,
\end{equation}
where, for an input $\mathbf{x}\in \mathds{R}^{M}$, its
neighborhood $\mathbf{Y(x)}\in \mathds{R}^{M\times K}$ is searched
from the training samples according to prior knowledge or manual
labeling. For simplicity, we assume that each data point has $K$
neighbors, and $\mathbf{c}_i\in\mathds{R}^{N}$ denotes the optimal
code for $\mathbf{y}_i(\mathbf{x})$.

To bridge the connection between the objective variants $\mathbf{a}$ and $\mathbf{c}_i$, it is possible to assume that $\mathbf{a}$
could be denotede as a linear combination of $\{\mathbf{c}_i\}_{i=1}^{K}$. Mathematically,
\begin{equation}
\label{equ:12}
\mathbf{a} =\sum_{\substack{\mathbf{c}_i\in \mathbf{C}}} w_i\mathbf{c}_i,
\end{equation}
where $w_i$ is the representation coefficient between $\mathbf{a}$ and $\mathbf{c}_i$. The calculation of $w_i$ is a challenging and key step which has been studied in many works. For example, Roweis and Saul~\cite{Roweis2000} defined $\mathbf{w}$ as the reconstruction coefficients over the nearby points in the original space. However, the approach is not suitable for our case since we aim to denote $\mathbf{c}_i$ with $\mathbf{a}$ but vice versa.

Motivated by a biological experiment of Ohki~\cite{Ohki2005} as discussed in Section~\ref{sec:1}, we present a simple but effective method to solve the problem by directly replacing $\mathbf{c}_i$ with $\mathbf{a}$. It is based on an observation (\figurename~\ref{fig:1}) that the representation of $\mathbf{x}$ also can approximate the representation of $\mathbf{y}_i$, i.e.,

\begin{equation*}
\|\mathbf{y}_i-\mathbf{Dc}_i\|_2^2 \leq
\|\mathbf{y}_i-\mathbf{Da}\|_2^2 \leq
\|\mathbf{y}_i-\mathbf{D\bar{a}}\|_2^2,
\end{equation*}
where $\bar{\mathbf{a}}$ denotes the representation of the point which is not close to $\mathbf{y}_i$.

Thus, the proposed objective function is as follows:
\begin{equation}
\label{equ:13}
  %\mathop{\mathrm{argmin}}_{\mathbf{a}}
    \mathrm{E}(\mathbf{x}, \mathbf{Y(x)}, \mathbf{a}) =(1-\gamma)\|\mathbf{x}-\mathbf{Da}\|_2^2
    +\gamma\frac{1}{K}\sum_{\substack{\mathbf{y}_i(\mathbf{x})\in \mathbf{Y(x)}}}\|\mathbf{y}_i-\mathbf{Da}\|_2^2 + \lambda \| \mathbf{a} \|_p,
\end{equation}

where $0\leqslant\gamma\leqslant1$ balances the importance between
the testing image $\mathbf{x}$ and its neighborhood
$\mathbf{Y(x)}$. The second term, which measures the contribution
of locality, can largely improve the robustness of \textbf{a}. If
$\mathbf{x}$ is corrupted by noise or occluded by disguise, a
larger $\gamma$ will yield better recognition results.

On the other hand, the locality constraint in (\ref{equ:13}) is a simplified model of the property that similar inputs having similar codes. We think this might be a new interesting way to learn local consistency.

Consider the recent findings, i.e., $\ell^1$-norm based sparsity cannot bring a higher recognition accuracy and better robustness for
facial data than $\ell^2$-norm based methods~\cite{Shi2011-recognition,Zhang2011-Sparse}, we simplify our objective
function (\ref{equ:13}) as follows:

\begin{equation}
\label{equ:14}
\mathrm{E}(\mathbf{x}, \mathbf{Y(x)}, \mathbf{a})
= (1-\gamma)\|\mathbf{x}-\mathbf{Da}\|_2^2
+\gamma\frac{1}{K}\sum_{\substack{\mathbf{y}_i(\mathbf{x})\in
\mathbf{Y(x)}}}\|\mathbf{y}_i(\mathbf{x})-\mathbf{Da}\|_2^2 +
\lambda \| \mathbf{a} \|_2^2.
\end{equation}

Clearly, (\ref{equ:14}) achieves the minimum when its derivative
with respect to $\mathbf{a}$ is zero. Hence, the optimal solution
is
\begin{equation}
\label{equ:16}
\mathbf{a^*}=(\mathbf{D}^T\mathbf{D}+\lambda \cdot
\mathbf{I})^{-1}\mathbf{D}^T\left[(1-\gamma)\mathbf{x}+\gamma\frac{1}{K}
\sum_{\substack{\mathbf{y}_i(\mathbf{x})\in
\mathbf{Y(x)}}}\mathbf{y}_i(\mathbf{x})\right].
\end{equation}

Let $\mathbf{P}=(\mathbf{D}^T\mathbf{D}+\lambda \cdot \mathbf{I})^{-1}\mathbf{D}^T$ whose calculation requires
re-formulating the psuedo-inverse, it can be
calculated in advance and only once as it is only dependent on
training data $\mathbf{D}$.

Given a testing image $\mathbf{x}$, the first step is to determine
its neighborhood $\mathbf{Y(x)}$ from the training set according
to prior knowledge, or manual labeling, etc. In practical
applications, there are two widely-used variations for finding the
neighborhood:
\begin{enumerate}
\item $\epsilon$-ball method: The training sample $\mathbf{d}_i$
is a neighbor of the testing image $\mathbf{x}$ if
$\|\mathbf{d}_i-\mathbf{x}\|_2<\epsilon$, where $\epsilon >0$ is a
constant.
 \item $K$-nearest neighbors (\emph{K}-NN) searching: The training sample $\mathbf{d}_i$ is a neighbor of $\mathbf{x}$, if $\mathbf{d}_i$ is among the $K$-nearest neighbors of $\mathbf{x}$, where $K>0$ can be specified as a constant or determined adaptively.
\end{enumerate}

Once the neighborhood of the testing image $\mathbf{x}$ is
obtained, LCCR just simply projects $\mathbf{x}$ and its
neighborhood $\mathbf{Y(x)}$ onto space $\mathbf{P}$ via
(\ref{equ:16}). In addition, the matrix form of LCCR is easily derived, which can used in batch
prediction.
\begin{equation*}
\label{equ:17}
    \mathbf{A^*}=(\mathbf{D}^T\mathbf{D}+\lambda \cdot \mathbf{I})^{-1}\mathbf{D}^T\left[(1-\gamma)\mathbf{X}+\gamma\frac{1}{K} \sum_{\substack{i=1}}^{K}\mathbf{Y}_i(\mathbf{X})\right],
\end{equation*}
where the columns of $\mathbf{X}\in \mathds{R}^{M\times J}$ are the testing images whose codes are stored in $\mathbf{A^*}\in \mathds{R}^{N\times J}$, and $\mathbf{Y}_i(\mathbf{X})\in \mathds{R}^{M\times J}$ denotes the collection of $i$th-nearest neighbor of $\mathbf{X}$.

The proposed LCCR algorithm is summarized in Algorithm~\ref{algorithm1}, and an overview is illustrated in \figurename~\ref{fig:2}.
\begin{algorithm}[ht]
    \caption{Face Recognition using Locality-Constrained Collaborative Representation (LCCR)}
    \label{algorithm1}
    \begin{algorithmic}[1]
    \REQUIRE
    A matrix of training samples $\mathbf{D}=[\mathbf{d}_1,\mathbf{d}_2,\ldots,\mathbf{d}_N]\in\mathds{R}^{M\times N}$ which are sorted according to the label of $\mathbf{d}_i$, $1\leq i \leq N$; A testing image $\mathbf{x}\in \mathds{R}^M$; The balancing factors $\lambda\geq0$, $0\leqslant\gamma\leqslant1$, and the size of neighborhood $K$.
    \STATE Normalize the columns of $\mathbf{D}$ and $\mathbf{x}$ to have a unit $\ell^2$-norm, respectively.
    \STATE Calculate the projection matrix $\mathbf{P}=(\mathbf{D}^T\mathbf{D}+\lambda \cdot \mathbf{I})^{-1}\mathbf{D}^T$ and store it.
    \STATE Find the neighborhood $\mathbf{Y(x)}$=\{$\mathbf{y}_1(\mathbf{x})$, $\mathbf{y}_2(\mathbf{x})$, $\ldots$, $\mathbf{y}_K(\mathbf{x})$\} for $\mathbf{x}$ from the training samples $\mathbf{D}$.
    \STATE Code $\mathbf{x}$ over $\mathbf{D}$ via
            \begin{equation*}
                \mathbf{a^*}=\mathbf{P}\cdot\left[(1-\gamma)\mathbf{x}+\gamma\frac{1}{K} \sum_{\substack{\mathbf{y}_i(\mathbf{x})\in \mathbf{Y(x)}}}\mathbf{y}_i(\mathbf{x})\right].
            \end{equation*}
    \STATE Compute the regularized residuals over all classes by
            \begin{equation*}
                r_i(\mathbf{x})=\frac{\|\mathbf{x}-\mathbf{D}*\delta_i(\mathbf{a^*})\|_2}{\|\delta_i(\mathbf{a^*})\|_2},
            \end{equation*}
            where $i$ denotes the index of class.
    \ENSURE $\mathrm{identity}(\mathbf{x})=\mathop{\mathrm{argmin}}_{i}\{r_i(\mathbf{x})\}$.
    \end{algorithmic}
\end{algorithm}

\subsection{Discussions}

\begin{figure*}[t]
\subfigure[]{\label{fig:add1.a}\centering\includegraphics[width=0.255\textwidth]{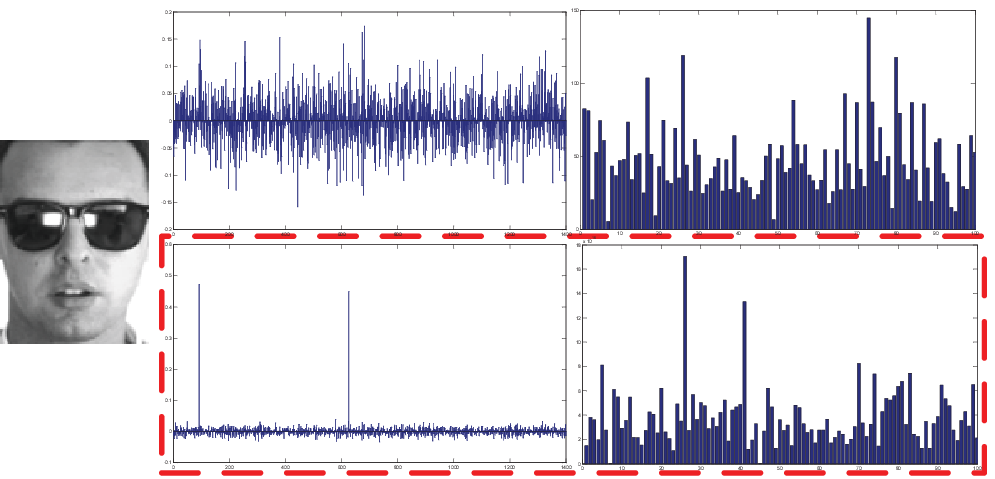}}\hspace{0.1cm}
\subfigure[]{\label{fig:add1.b}\centering\includegraphics[width=0.22\textwidth]{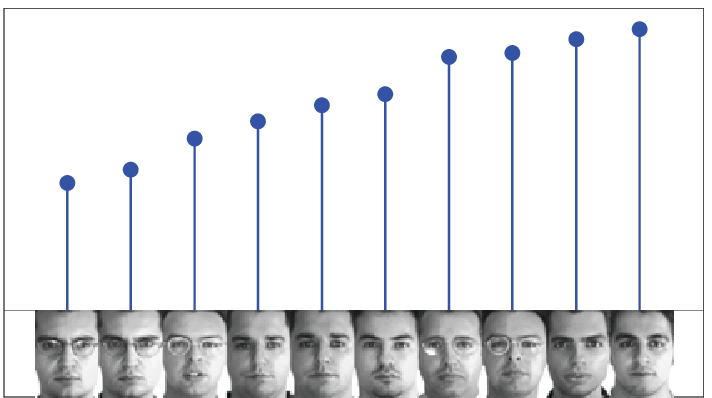}}\hspace{0.1cm}
\subfigure[]{\label{fig:add1.c}\centering\includegraphics[width=0.255\textwidth]{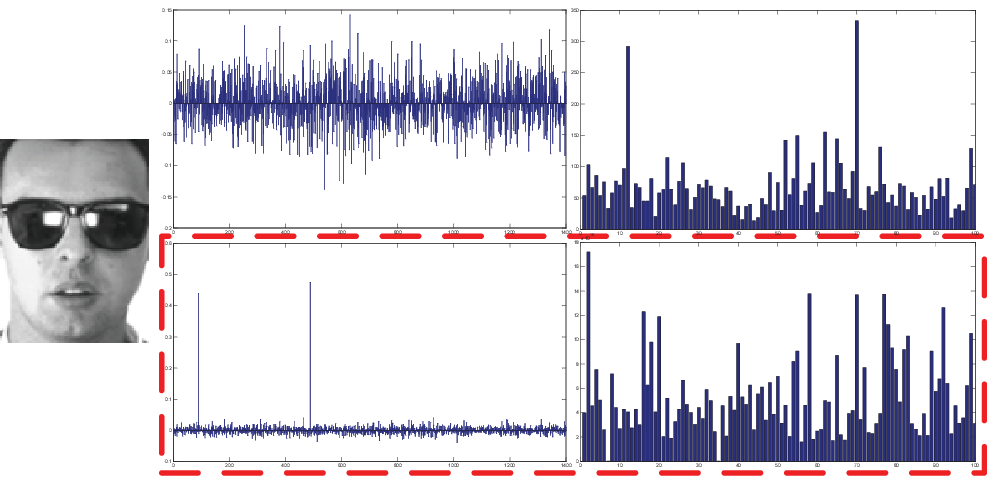}}\hspace{0.1cm}
\subfigure[]{\label{fig:add1.d}\centering\includegraphics[width=0.2165\textwidth]{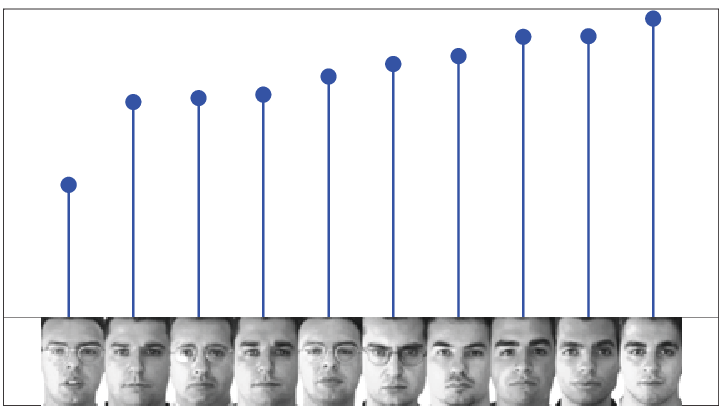}}
\caption{The effectiveness of the proposed model. (a) A testing face disguised by sunglass comes from the 7th subject of AR database. The figures (in the red rectangle) in the second row are the coefficients and residual of the input learned by LCCR ($\lambda=0.005$, $\gamma=0.9$, and $k=2$); the figures in the first row are the results of CRC-RLS~\cite{Zhang2011-Sparse} ($\lambda=0.001$ for the best accuracy). (b) The 10 nearest neighbors of the input in terms of cityblock distance (Y-axis). (c) and (d) are the results of another testing sample from the same individual.}
\label{fig:add1}
\end{figure*}

From the algorithm, it is easy to see that the performance of LCCR is positively correlated with that of $K$-NN searching method. Thus, it is possible to assume that LCCR will be failed if $K$-NN cannot find the correct neighbors for the testing sample. Here, we give a real example (\figurename\ref{fig:add1}) to illustrate that LCCR would largely avoid such situations from happening. In the example, the classification accuracy of LCCR is about 94\% by using 600 AR images with sunglasses as testing image and 1400 clean ones as training samples.

\figurename~\ref{fig:add1.a} demonstrates the coefficients and residual of LCCR and CRC-RLS. We can see that the two methods correctly predicted the identity of the input, while $K$-NN searching could not find the correct neighbors (see \figurename~\ref{fig:add1.b}). It illustrates that LCCR could work well even though $K$-NN is failed to get the results. \figurename~\ref{fig:add1.c} and \figurename~\ref{fig:add1.d} illustrate another possible case. That is, CRC-RLS fails to get the correct identity of the input while the nearest neighbor cames from the 7th individual, and LCCR successfully obtains the the correct identify.

\subsection{Computational Complexity Analysis}

The computational complexity of LCCR consists of two parts for offline and online computation, respectively. Suppose the dictionary $\mathbf{D}$ contains $n$ samples with $m$ dimensionality, LCCR takes $O(mn^2+n^3)$ to compute the projection matrix $(\mathbf{D}^T\mathbf{D}+\lambda \mathbf{I})^{-1}\mathbf{D}^{T}$ and $O(mn)$ to store it.

For each querying sample $\mathbf{y}$, LCCR needs $O(mn)$ to search the \emph{K}-nearest neighbors of $\mathbf{y}$ from $\mathbf{D}$. After that, the algorithm projects $\mathbf{y}$ into another space via (\ref{equ:16}) in $O(m^2n)$. Thus, the computational complexity of encoding LCCR is $O(m^2n)$ for each unknown sample. Note that, the computational complexity of LCCR is same with that of LRC~\cite{Shi2011-recognition,Naseem2010-Linear} and CRC-RLS~\cite{Zhang2011-Sparse}, and it is more competitive than SRC~\cite{Wright2009-Robust} even though the fastest $\ell^1$-solver is used. For example, SRC takes $O(t_1 m^2 n+t_1 mn^2)$ to code each sample over $\mathbf{D}$ when Homotopy optimizer~\cite{Osborne2000} is adopted to get the sparsest solution, where Homotopy optimizer is one of the fastest $\ell^1$-minimization algorithm according to~\cite{Yang2010-l1-minimization} and $t$ denotes the number of iterations of Homotopy algorithm. From the above analysis, it is easy to find that a medium-sized data set will bring up the scalability issues with the models. To address the problem, a potential choice is to perform dimension reduction or sampling techniques to reduce the size of problem in practical application as did in~\cite{Peng2013}.

\section{Experimental Verification and Analysis}
\label{sec:4}

In this section, we report the performance of LCCR over four publicly-accessed facial databases, i.e., AR~\cite{Martinez1998},
ORL~\cite{Samaria1994}, the Extended Yale database B~\cite{Georghiades2001}, and Multi-PIE~\cite{Gross2010}. We examine the recognition results of the proposed algorithm with respect to 1) discrimination, 2) robustness to corruptions, 3) and robustness to occlusions.

\subsection{Experimental Configuration}
\label{sec:4.1}

We compared the classification results of LCCR with four linear coding models (SRC~\cite{Wright2009-Robust}, CESR~\cite{He2011-Maximum},
LRC~\cite{Shi2011-recognition,Naseem2010-Linear} and CRC-RLS~\cite{Zhang2011-Sparse}) and a subspace learning algorithm (LPP~\cite{He2003-Locality}) with the nearest neighbors classifier (1NN). Moreover, we also reported the results of linear SVM~\cite{Fan2008} over the original inputs. Note that, SRC, CESR, LRC, CRC-RLS and LCCR directly code each testing sample over training data without usability of dictionary learning method, and get classification result by finding which subject produces the minimum reconstruction error. In these models, only LCCR incorporates locality based pairwise distance into coding scheme. For a comprehensive comparison, we report the performance of LCCR with five basic distance metrics, i.e., Euclidean distance ($\ell^2$-distance), Seuclidean distance (standardized Euclidean distance), Cosine distance (the cosine of the angle between two points), Cityblock distance
($\ell^1$-distance), and Spearman distance.

For computational efficiency, as did in~\cite{Wright2009-Robust,Zhang2011-Sparse}, we performed Eigenface~\cite{Turk1991} to reduce the dimensionality of data set throughout the experiments. Moreover, SRC requires the dictionary $\mathbf{D}$ to be an under-determined matrix, and Shi et al.~\cite{Shi2011-recognition} claimed that their model (named as LRC in~\cite{Naseem2010-Linear}) will achieve competitive results when $\mathbf{D}$ is over-determined. For a extensive comparison, we investigate the performance of the tested methods except SRC over two cases.

We solved the $\ell^1$-minimization problem in SRC by using the CVX~\cite{Grant2008}, a package for solving convex optimization problems, and got the results of LRC, CRC-RLS and CESR by using the source codes from the homepages of the authors. All experiments are carried out using Matlab 32bit on a 2.5GHz machine with 2.00 GB RAM.

\begin{figure*}[t]
\subfigure []{\label{fig:3.a}\centering\includegraphics[width=0.325\textwidth]{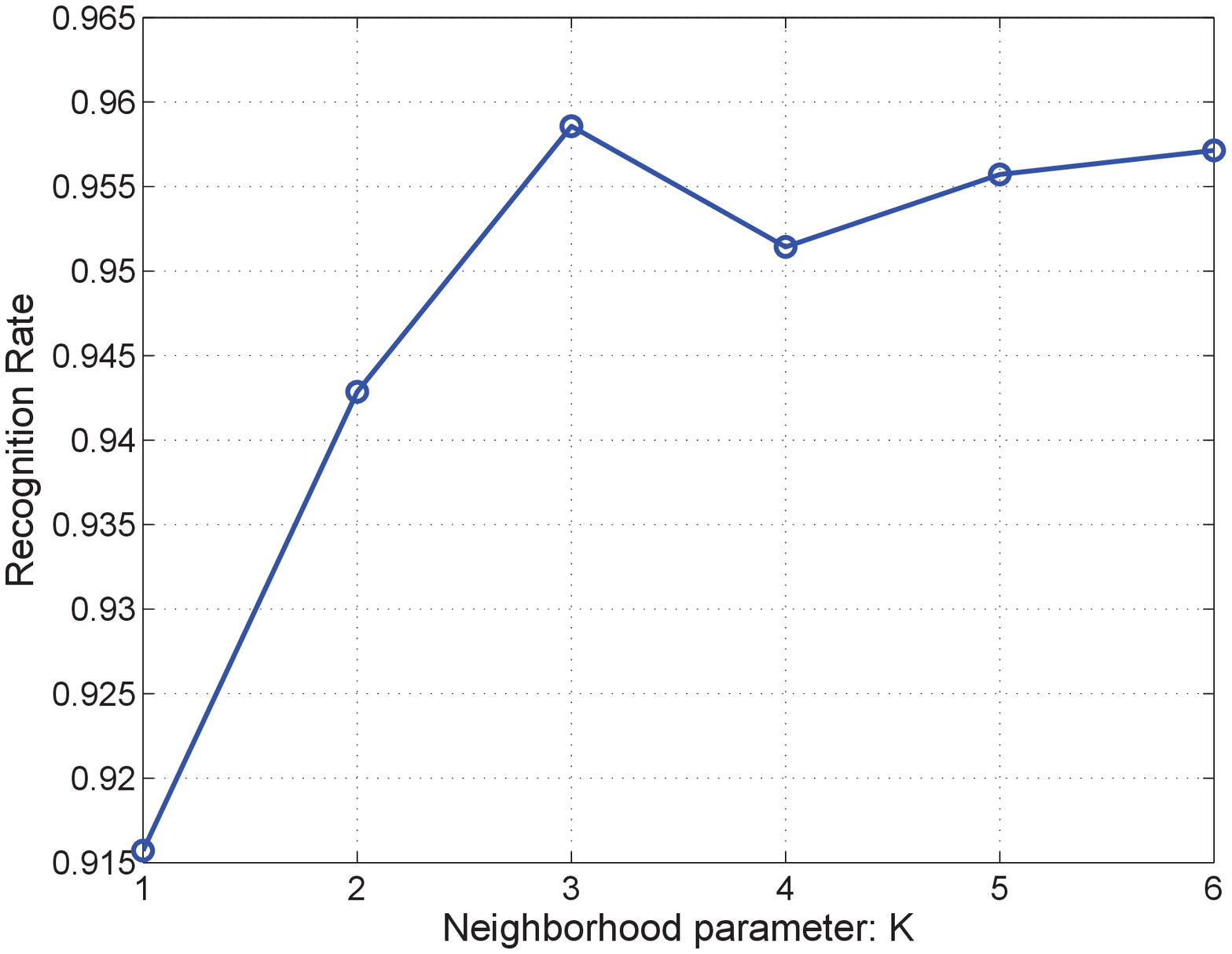}}
\subfigure []{\label{fig:3.b}\centering\includegraphics[width=0.325\textwidth]{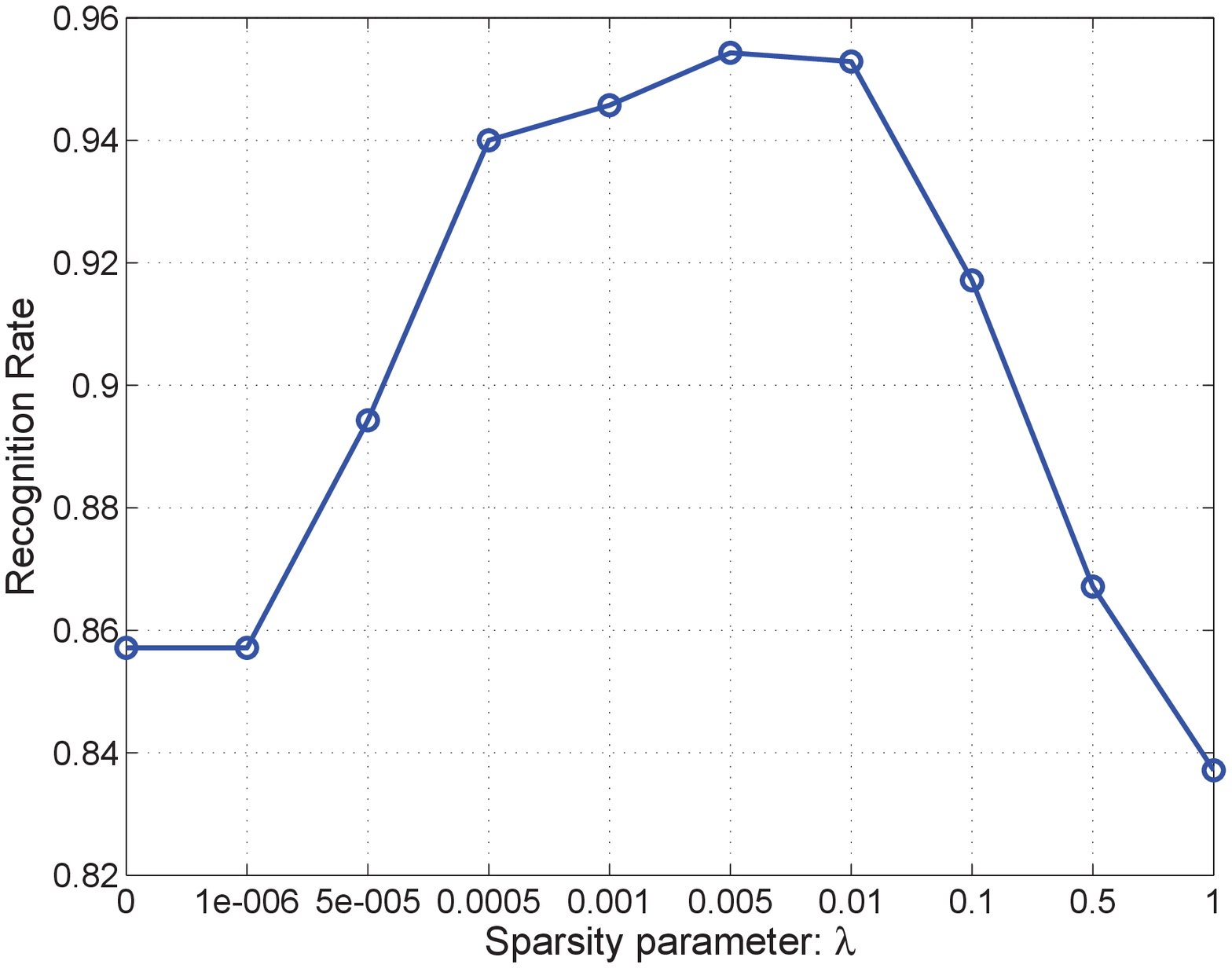}}
\subfigure []{\label{fig:3.c}\centering\includegraphics[width=0.325\textwidth]{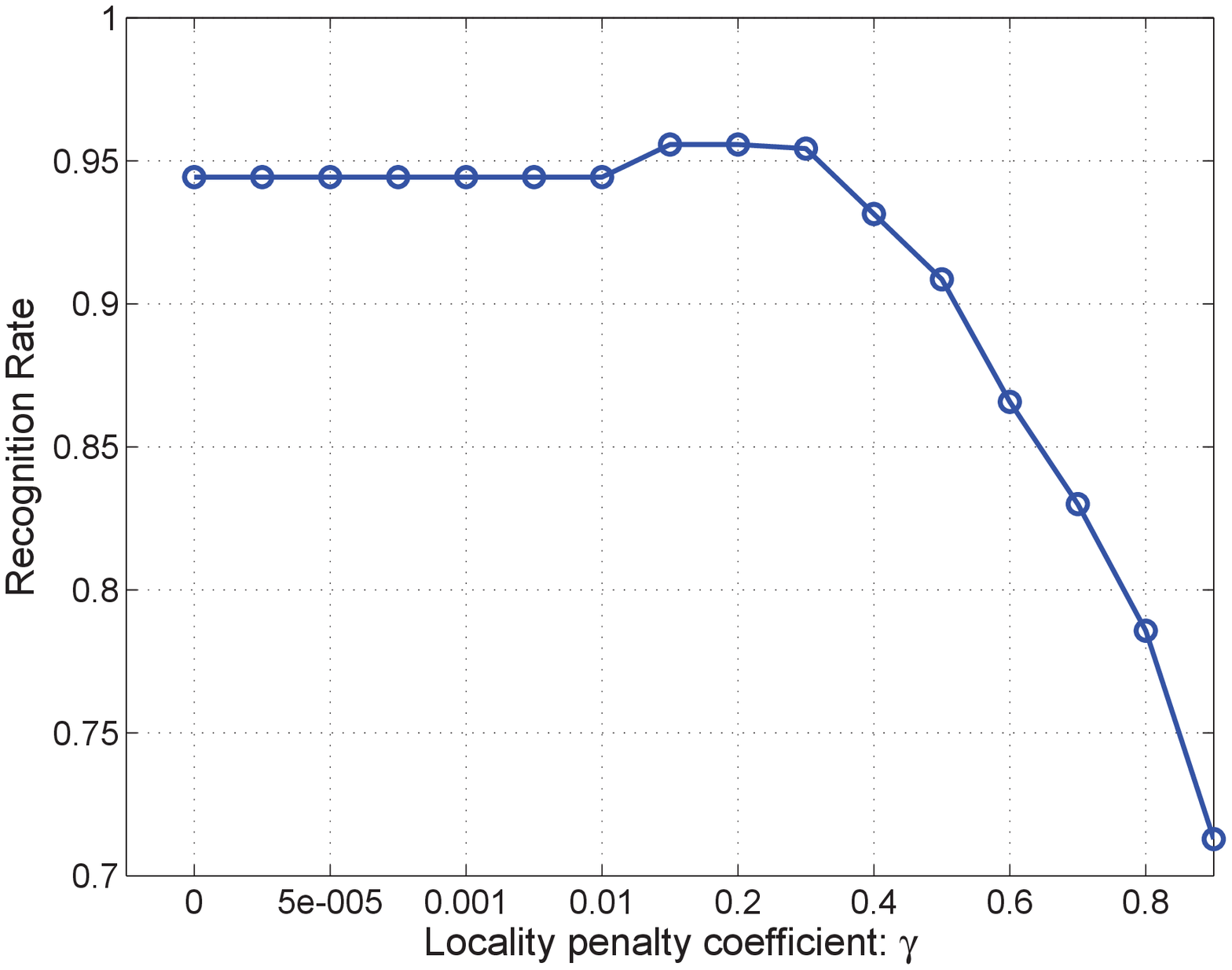}}
\caption{Recognition accuracy of LCCR using Cityblock distance on a subset of AR database with dimensionality 2580. (a) The recognition rates versus the variation of the neighborhood parameter $K$, where $\lambda=0.005$ and $\gamma=0.2$. (b) The recognition rates versus the variation of the sparsity parameter $\lambda$, where $K=5$ and $\gamma=0.2$. (c) The recognition rates versus the variation of the locality constrained coefficient $\gamma$, where $K=3$ and $\lambda=0.005$.}
\label{fig:3}
\end{figure*}

Parameter determination is a big challenge in pattern recognition and computer vision. As did in \cite{Cheng2010-Learning,Peng2012}, we report the
best classification results of all tested methods under different parameter configurations. The value range used to find the best values for LCCR can be inferred from \figurename~\ref{fig:3}, and these possible values of $\lambda$ also are tested for SRC and CRC-RLS. In all tests, we randomly split each data set into two parts for training and testing, and compare the performance of the algorithms using the same partition
to avoid the difference in data sets.

\subsection{Recognition on Clean Images}
\label{sec:4.3}

In this sub-section, we examine the performance of 7 competing methods over 4 clean facial data sets. Here, clean image means an image without occlusion or corruption, just with variations in illumination, pose, expression, etc.

(1) \textbf{ORL database}~\cite{Samaria1994} consists of 400 different images of 40 individuals. For each person, there are 10 images with the
variation in lighting, facial expression and facial details (with or without glasses). For computational efficiency, we cropped all
ORL images from $112\times 96$ to $56\times48$, and randomly selected 5 images from each subject for training and used the
remaining 5 images for testing.

\tablename~\ref{tab:1} reports the classification accuracy of the tested algorithms over various dimensionality. Note that, the Eigenface with 200D retains $100\%$ energy of the cropped data, which makes the investigated methods achieve the same rates over 2688D. From the results, LCCRs outperform the other algorithms, and the best results are achieved when Cityblock distance is used to search the nearest neighbors. Moreover, we can find that all the algorithms achieve a higher recognition rate in the original space except LRC. One possible reason is that the cropped operation degrades the performance of LRC, another reason may attribute to the used classier. Moreover, we have found that if another nearest subspace classifier~\cite{Wright2009-Robust} is adopted with linear regression based representation, the accuracy of LRC is slightly decreased from 89\% to 88.00\% over the original data and from 91\% to 90\% with 120D.

\begin{table}[t]
\caption{The Maximal Recognition Accuracy of Competing Algorithms on the ORL Database.}
\label{tab:1}
\centering
\begin{scriptsize}
\begin{tabular}{|c|c|c|c|c|}
\hline
\bfseries Dim &  \bfseries 54 & \bfseries 120 & \bfseries 200 & \bfseries 2688\\
\hline\hline
SVM~\cite{Fan2008}   & 90.00\% & 92.50\% & 93.50\% & 93.00\%\\
LPP~\cite{He2003-Locality}  & 86.00\% & 86.50\% & 86.50\% & 86.50\%\\
SRC~\cite{Wright2009-Robust}   & 92.00\% & 96.50\% & 86.00\% & - \\
CESR~\cite{He2011-Maximum}  & 89.50\%  & 88.50\%  & 89.00\%  & 97.50\%\\
LRC~\cite{Shi2011-recognition,Naseem2010-Linear} & 92.50\%    & 91.00\% & 89.00\% & 89.00\%\\
CRC-RLS~\cite{Zhang2011-Sparse}  & 94.50\% & 94.00\% & 94.50\% & 95.00\%\\
\hline
LCCR + Cityblock & \bfseries97.50\% & \bfseries97.50\% & \bfseries98.00\% & \bfseries98.00\%\\
LCCR + Seuclidean & 96.00\% & 96.50\% & 96.00\% & 96.50\%\\
LCCR + Euclidean & 96.00\% & 96.00\% & 96.50\% & 96.50\%\\
LCCR + Cosine & 96.00\% & 96.50\% & 96.50\% & 96.50\%\\
LCCR + Spearman & 96.00\% & 96.00\% & 96.00\% & 96.00\%\\
\hline
\end{tabular}
\end{scriptsize}
\end{table}

%\subsubsection{AR database}
%\label{sec:4.3.1}

(2) \textbf{AR database}~\cite{Martinez1998} includes over 4000 face images of 126 people (70 male and 56 female) which vary in expression, illumination and disguise (wearing sunglasses or scarves). Each subject has 26 images consisting of 14 clean images, 6 images with sunglasses and 6 images with scarves. As did in~\cite{Wright2009-Robust, Zhang2011-Sparse}, a subset that contains 1400 normal faces randomly selected from 50 male subjects and 50 female subjects, is used in our experiment. For each subject, we randomly permute the 14 images and take the first half for training and the rest for testing. Limited by the
computational capabilities, as in \cite{Zhang2011-Sparse}, we crop all images from original $165\times 120$ to $60\times 43$ (2580D) and convert it to gray scale.
%From \tablename~\ref{tab:2}, we can see that LCCRs are superior to SVM, SRC, CESR, LRC and CRC-RLS with respect to all examined dimensionality.

\begin{table}[t]
\caption{The Maximal Recognition Accuracy of Competing Algorithms on the AR Database.}
\label{tab:2}
\centering
\begin{scriptsize}
\begin{tabular}{|c|c|c|c|c|}
\hline
\bfseries Dim &  \bfseries 54 & \bfseries 120 & \bfseries 300 & \bfseries 2580\\
\hline\hline
SVM~\cite{Fan2008}   & 73.43\% & 81.00\% & 82.00\% & 83.14\%\\
LPP~\cite{He2003-Locality} & 39.29\% & 43.57\% & 53.86\% & 53.86\%\\
SRC~\cite{Wright2009-Robust}   & 81.71\% & 88.71\% & 90.29\% & -\\
CESR~\cite{He2011-Maximum} & 74.00\% & 81.43\% & 84.57\% & 84.57\%\\
LRC~\cite{Shi2011-recognition,Naseem2010-Linear} & 80.57\%  & 90.14\% & 93.57\% & 82.29\%\\
CRC-RLS~\cite{Zhang2011-Sparse}   & 80.57\% & 90.43\% & 94.00\% & 94.43\%\\
\hline
LCCR + Cityblock & \bfseries86.14\% & \bfseries92.71\% & \bfseries95.14\% & \bfseries95.86\%\\
LCCR + Seuclidean & 85.00\% & 91.86\% & 94.43\% & 95.43\%\\
LCCR + Euclidean & 84.00\% & 91.29\% & 94.14\% & 94.86\%\\
LCCR + Cosine & 83.43\% & 90.86\% & 94.00\% & 94.57\%\\
LCCR + Spearman & 84.71\% & 90.71\% & 94.14\% & 94.43\%\\
\hline
\end{tabular}
\end{scriptsize}
\end{table}

%\subsubsection{Extended Yale B database}
%\label{sec:4.3.3}

(3) \textbf{Extended Yale B database}~\cite{Georghiades2001} contains 2414 frontal-face images with size $192\times 168$ over 38 subjects, as did in~\cite{Wright2009-Robust,Zhang2011-Sparse}, we carried out the experiments on the cropped and normalized images of size $54\times 48$. For each
subject (about 64 images per subject), we randomly split the images into two parts with equal size, one for training, and the other for testing. Similar to the above experimental configuration, we calculated the recognition rates over dimensionality 54, 120 and 300 using Eigenface, and 2592D in the original data space. \tablename~\ref{tab:3} show that LCCRs again outperform its counterparts across various spaces, especially when the Spearman distance is used to determine the neighborhood of testing samples.

\begin{table}[t]
\caption{The Maximal Recognition Accuracy of Competing Algorithms on the Extended Yale B Database.}
\label{tab:3}
\centering
\begin{scriptsize}
\begin{tabular}{|c|c|c|c|c|}
\hline
\bfseries Dim &  \bfseries 54 & \bfseries 120 & \bfseries 300 & \bfseries 2592\\
\hline\hline
SVM~\cite{Fan2008}   & 84.52\%  & 92.72\% & 95.28\% & 95.45\%\\
LPP~\cite{He2003-Locality} & 35.93\% & 54.55\% & 70.78\% & 75.66\%\\
SRC~\cite{Wright2009-Robust}   & 93.71\%   & 95.12\% & 96.44\% & -\\
CESR~\cite{He2011-Maximum} & 92.30\% & 94.95\% & 95.53\% & 96.11\%\\
LRC~\cite{Shi2011-recognition,Naseem2010-Linear} & 92.88\% & 95.61\% & 97.85\% & 90.48\%\\
CRC-RLS~\cite{Zhang2011-Sparse}  & 92.96\% & 95.69\% & 97.90\% & 98.26\%\\
\hline
LCCR + Cityblock & 93.21\% & 96.03\% & 97.93\% & 98.34\%\\
LCCR + Seuclidean & 93.21\% & 95.70\% & 97.93\% & 98.34\%\\
LCCR + Euclidean & 93.21\% & 95.70\% & 97.93\% & 98.34\%\\
LCCR + Cosine & 93.46\% & 95.78\% & 97.93\% & 98.59\%\\
LCCR + Spearman & \bfseries97.02\% & \bfseries98.18\% & \bfseries99.10\% & \bfseries99.59\%\\
\hline
\end{tabular}
\end{scriptsize}
\end{table}

%\subsubsection{Multi PIE database}
%\label{sec:4.3.4a}

(4) \textbf{Multi PIE database} (MPIE) ~\cite{Gross2010} contains the images of 337 subjects captured in 4 sessions with simultaneous variations in pose, expression and illumination.  As did in~\cite{Zhang2011-Sparse}, we used all the images in the first session as training data and the images belonging to the first 250 subjects in the other sessions as testing data. All images are cropped from $100\times 82$ to $50\times 41$.

\begin{table*}[t]
\caption{The Maximal Recognition Accuracy of Competing Algorithms on the Multi PIE database.}
\label{tab:4}
\centering
\begin{scriptsize}
\begin{tabular}{|c|c|c|c|c|c|c|}
\hline
\bfseries Dim & \multicolumn{3}{c|}{\multirow{1}{*}{\bfseries 300}} & \multicolumn{3}{c|}{\multirow{1}{*}{\bfseries 2050}}\\
\hline
Dataset & MPIE-S2 & MPIE-S3 & MPIE-S4 & MPIE-S2 & MPIE-S3 & MPIE-S4\\
\hline\hline
SVM~\cite{Fan2008}  &  91.33\% & 85.13\% & 89.20\% & 91.45\% & 85.75\% & 89.43\%\\
LPP~\cite{He2003-Locality} & 40.12\% & 27.44\% & 31.20\% & 31.49\% & 31.00\% & 31.49\%\\
SRC~\cite{Wright2009-Robust}  & 93.13\% & 90.60\% & 94.10\% & - & - & -                     \\
CESR~\cite{He2011-Maximum} & 92.41\% & 87.38\% & 91.94\% & 94.46\% & 92.06\% & \bfseries 96.17\%\\
LRC~\cite{Shi2011-recognition,Naseem2010-Linear} & 94.64\% & 89.88\% & 93.37\% & 83.19\% & 70.25\% & 75.03\%\\
CRC-RLS~\cite{Zhang2011-Sparse}  & 94.88\% & 89.88\% & 93.60\% & 95.30\% & 90.56\% & 94.46\%\\
\hline
LCCR + Cityblock & \bfseries 95.36\% & 91.25\% & \bfseries 95.54\% & \bfseries 96.08\% & 91.94\% & 95.89\%\\
LCCR + Seuclidean & 95.06\% & 91.25\% & 94.51\% & 95.84\% & 91.56\% & 95.20\%\\
LCCR + Euclidean & 95.06\% & 91.31\% & 94.57\% & 95.84\% & 91.63\% & 95.14\%\\
LCCR + Cosine & 95.12\% & 90.88\% & 94.69\% & 95.78\% & 91.38\% & 95.14\%\\
LCCR + Spearman & 95.12\% & \bfseries 91.75\% & 94.40\% & 95.78\% & \bfseries92.19\% & 95.20\%\\
\hline
\end{tabular}
\end{scriptsize}
\end{table*}

From Tables~\ref{tab:1}-\ref{tab:4}, we draw the following conclusions:
\begin{enumerate}
  \item LCCRs generally outperforms SVM (original input), SRC (sparse representation), CESR (robust sparse representation), LRC (linear regression based model) and CRC-RLS (collaborative representation) over the tested cases.
  \item LCCRs perform better in a low-dimensional space than a high-dimensional ones. For example, on the Extended Yale B, the difference in accuracy between LCCR and CRC-RLS (the second best method) changed from $4.06\%$ (54D) to $2.49\%$ (120D) and to $1.17\%$ (300D). It again corroborates our claim that local consistency is helpful to improving the discrimination of data representation, since the low-dimensional data contain few information than higher one.
  \item CESR is more competitive in the original space at the cost of computing cost. For example, it outperforms the other models over MPIE-S4 in classification accuracy where its time cost about 11003.51 seconds, compared with 3104.82s of SRC, 54.59s of LRC, 54.79s of CRC-RLS and 59.82s of LCCR.
  \item SRC, LRC and CRC-RLS achieve the similar performance, and SRC is more competitive in the low-dimensional feature spaces. The results are consistent with the reports in~\cite{Zhang2011-Sparse}. For example, in the experiments of Zhang over MPIE-S2 with 300D, the accuracy scores of SRC and CRC-RLS are about 93.9\% and 94.1\%, respectively, comparing with 93.13\% and 94.88\% in our experiments. Moreover, CRC-RLS and LRC achieve similar recognition rates with the difference less than $1\%$ across various feature spaces.
\end{enumerate}

\subsection{Recognition on Partial Facial Features}
\label{sec:4.4}

\begin{figure}[t]
\centering
\subfigure []{\label{fig:6.a}\includegraphics[width=0.36\textwidth]{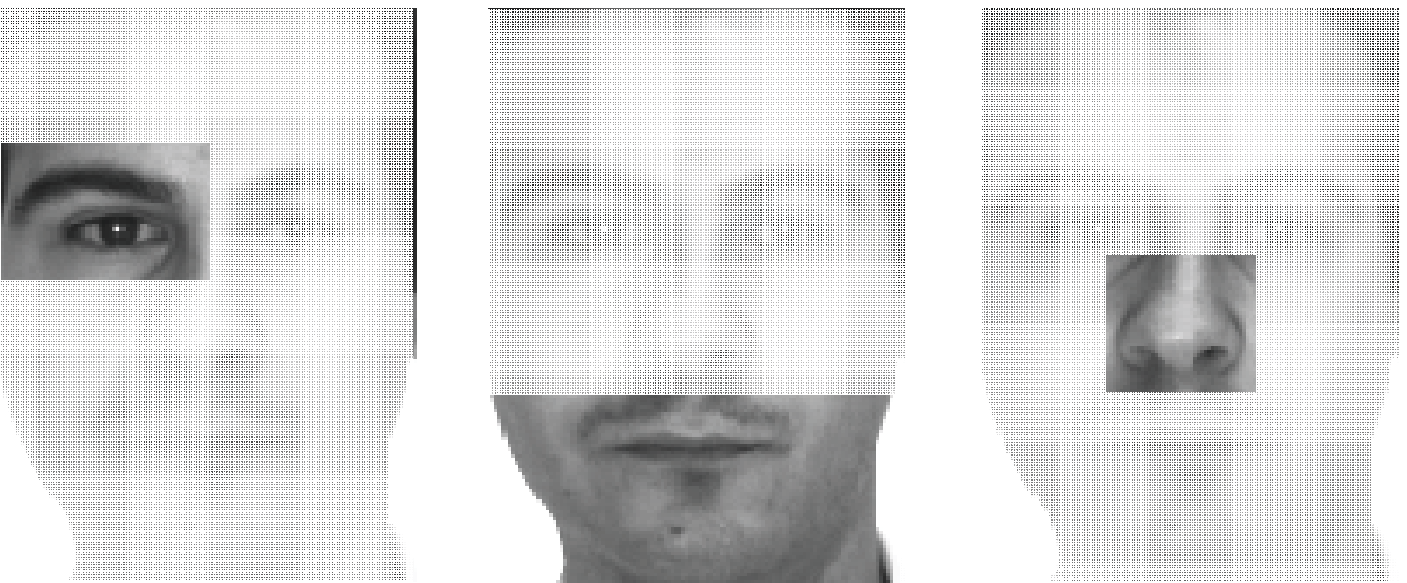}}
\subtable []{
\label{fig:6.b}
\begin{scriptsize}
\begin{tabular}{|c|c|c|c|}
\hline
\bfseries Features &  \bfseries Right Eye & \bfseries Mouth and Chin & \bfseries Nose \\
\hline
\bfseries Dim &  \bfseries 308 & \bfseries 798 & \bfseries 224\\
\hline\hline
SVM~\cite{Fan2008}   & 70.71\%  & 41.29\% & 37.14\% \\
LPP~\cite{He2003-Locality} & 58.57\% & 14.43\% & 53.71\%\\
SRC~\cite{Wright2009-Robust}   & 84.00\%   & 70.71\% & \bfseries 78.00\% \\
CESR~\cite{He2011-Maximum} & 81.57\% & 56.00\% & 70.43\%\\
LRC~\cite{Shi2011-recognition,Naseem2010-Linear} & 72.86\%    & 32.86\% & 70.57\% \\
CRC-RLS~\cite{Zhang2011-Sparse}  & 83.14\% & 73.86\% & 73.57\% \\
\hline
LCCR + Cityblock & \bfseries 86.86\% & \bfseries76.29\% & 75.86\% \\
LCCR + Seuclidean & 84.86\% & 75.57\% & 75.29\% \\
LCCR + Euclidean & 85.29\% & 75.00\% & 76.00\% \\
LCCR + Cosine & 84.43\% & 74.57\% & 75.29\%\\
LCCR + Spearman & 84.57\% & 75.86\% & 75.14\%\\
\hline
\end{tabular}\end{scriptsize}}
\caption{Recognition Accuracy with partial face features. (a) An example of the three features, right eye, mouth and chin, and nose from left to right. (b) The recognition rates of competing methods on the partial face features of the AR database.}
\label{fig:6}
\end{figure}

The ability to work on partial face features is very interesting since not all facial features play an equal role in recognition. Therefore, this ability has become an important metric in the face recognition researches~\cite{Savvides2006}. We examine the performance of the investigated methods using three partial facial features, i.e., right eye, nose, as well as mouth and chin, sheared from the clean AR faces with 2580D (as shown in \figurename~\ref{fig:6.a}). For each partial face feature, we generate a data set by randomly selecting 7 images per subject for training and the remaining 700 for testing. It should be noted that~\cite{Wright2009-Robust} conducted the similar experiment on Extended Yale B which includes less subjects, smaller irrelevant white
background, and more training samples per subject than our case.

\figurename~\ref{fig:6.b} shows that LCCRs achieve better recognition rates than SVM, SRC, LRC and CRC-RLS for right eye as well as mouth and chin, and the second best rates for the nose. Some works found that the most important feature is the eye, followed by the mouth, and then the nose~\cite{Sinha2006-Recognition}. We can see that the results for SVM, CRC-RLS and LCCR are consistent with the conclusions even though the
dominance of the mouth and chin over the nose is not very distinct.

\subsection{Face Recognition with Block Occlusions}
\label{sec:4.5}

\begin{figure*}[t]
\centering
\subfigure []{\label{fig:7.a}\includegraphics[width=0.056\textwidth]{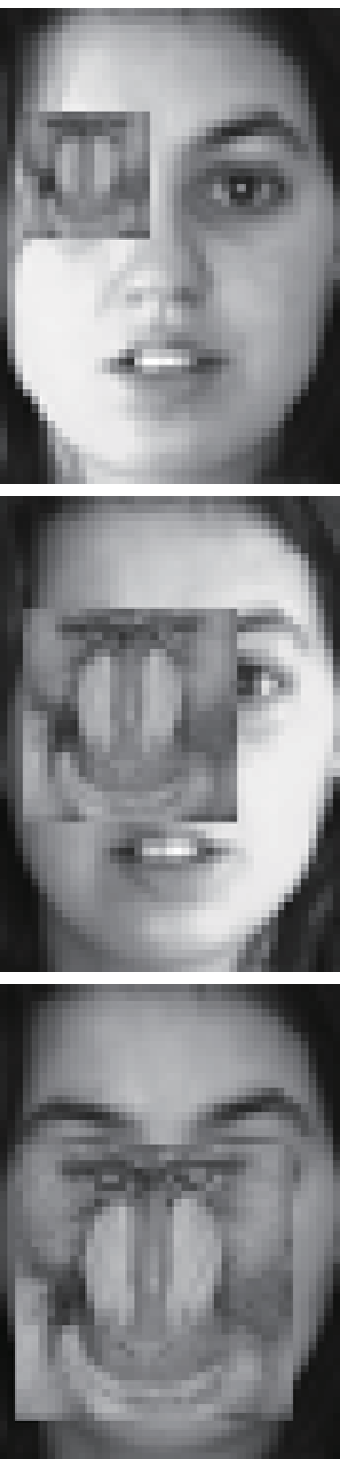}}\hspace{0.1cm}
\subfigure []{\label{fig:7.b}\includegraphics[width=0.43\textwidth]{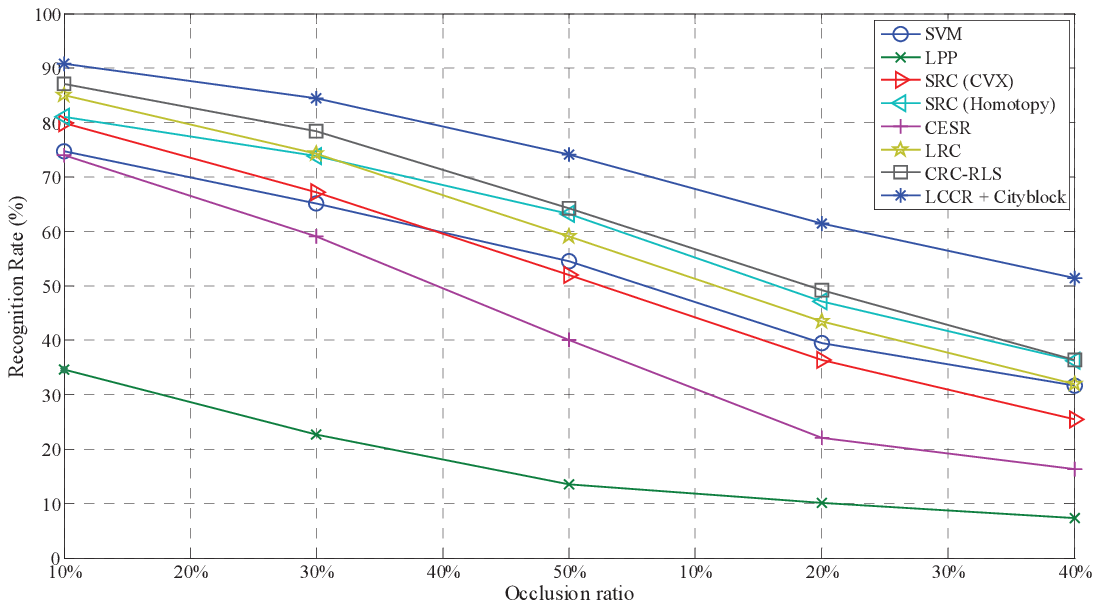}}\hspace{0.1cm}
\subfigure []{\label{fig:7.c}\includegraphics[width=0.43\textwidth]{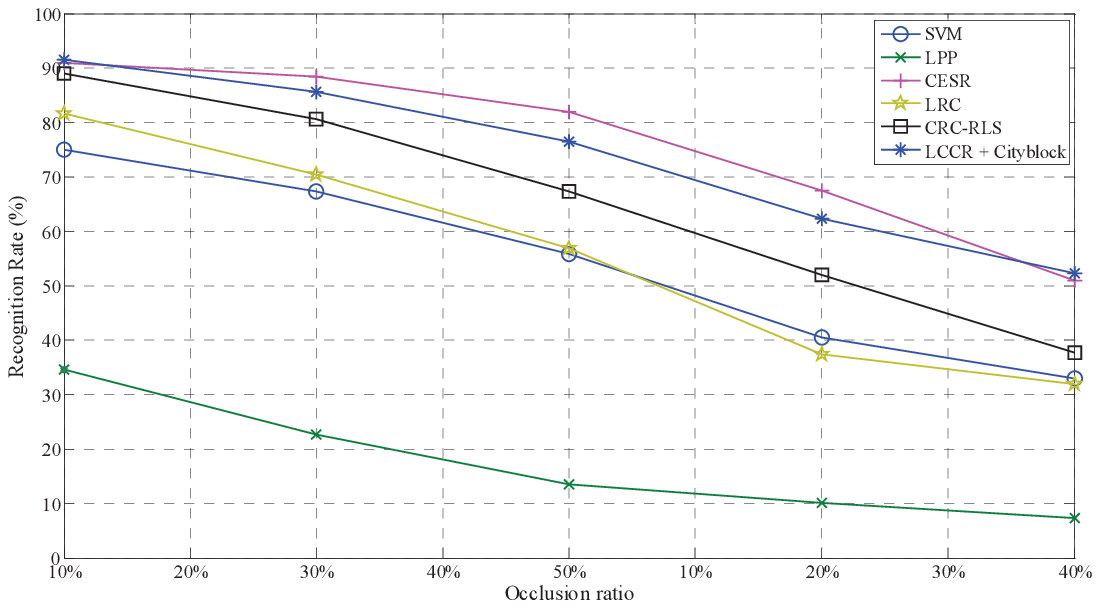}}\hspace{0.1cm}
\begin{scriptsize}
\subtable []{
\label{fig:7.d}
\begin{tabular}{|c||c|c|c|c|c| c|c|c|c|c|}
\hline
\bfseries Dim & \multicolumn{5}{c|}{\bfseries 300 (Eigenface)} & \multicolumn{5}{c|}{\bfseries 2580}\\
\hline
\bfseries Occlusion ratio &  \bfseries 10\% & \bfseries 20\% & \bfseries 30\% & \bfseries 40\% & \bfseries 50\% &  \bfseries 10\% & \bfseries 20\% & \bfseries 30\% & \bfseries 40\% & \bfseries 50\%\\
\hline\hline
LCCR + Cityblock  & 90.71\% & 84.43\% & 74.14\% & 61.43\% & 51.43\% & 91.57\% & 85.57\% & 76.43\% & 62.29\% & 52.29\%\\
LCCR + Seuclidean & 90.00\% & 80.14\% & 67.86\% & 53.71\% & 39.57\% & 91.00\% & 82.14\% & 70.57\% & 56.29\% & 41.71\%\\
LCCR + Euclidean & 88.86\% & 80.86\% & 68.14\% & 55.00\% & 43.29\% & 89.71\% & 81.86\% & 71.00\% & 57.00\% & 43.86\%\\
LCCR + Cosine & 88.57\% & 80.29\% & 69.29\% & 56.57\% & 44.00\% & 89.00\% & 82.43\% & 71.00\% & 57.29\% & 45.57\%\\
LCCR + Spearman & 88.29\% & 81.00\% & 70.00\% & 57.14\% & 43.29\% & 89.57\% & 83.00\% & 72.57\% & 59.43\% & 44.14\%\\
\hline
\end{tabular}}
\end{scriptsize}
\caption{Experiments on AR database with varying percent block
occlusion. (a) From top to bottom, the occlusion percents for test
images are, $10\%$, $30\%$, and $50\%$, respectively. (b) and (c)
are the recognition rates under different levels of block
occlusion on AR database with 300D (Eigenface) and 2580D,
respectively. (d) The recognition rates of LCCRs with 300D and 2580D.}
\label{fig:7}
\end{figure*}

To examine the robustness to block occlusion, similar to~\cite{Wright2009-Robust, Shi2011-recognition, Zhang2011-Sparse}, we get 700 testing
images by replacing a random block of each clean AR image with an irrelevant image (baboon) and use 700 clean images for training.
The occlusion ratio increases from 10\% to 50\%, as shown in \figurename~\ref{fig:7.a}. We investigate the classification accuracy of the methods across Eigenface space with 300D (\figurename~\ref{fig:7.b}) and cropped data space with 2580D (\figurename~\ref{fig:7.c}).

\figurename s~\ref{fig:7.b}-\ref{fig:7.d} show that LCCRs generally outperform the other models with considerable performance margins. Especially, with the increase of the occlusion ratio, the difference in recognition rates of LCCRs and the other methods becomes larger. For example, when the occlusion ratio is $50\%$, in 300 dimensional space, the accuracy of LCCR with Cityblock distance is about $19.7\%$ higher than SVM, about $44.1\%$ higher than LPP, about $26.0\%$ higher than SRC (CVX), about $35.1\%$ higher than CESR, about $19.6\%$ higher than LRC, and about $15.1\%$ higher than CRC-RLS. Note that, different $\ell^1$-solvers will lead to different results for SRC. For Example, if SRC adopts Homotopy algorithm~\cite{Osborne2000} to get the sparest solution, the recognition rate will increase from 25.43\% (with CVX) to 36.14\% such that the performance dominance decreases from $26\%$ to $15.3\%$. Moreover, CESR achieves the best results at the cost of computational cost when the original data is available and the occluded ratio ranges from 20\% to 40\%.

\begin{figure*}[t]
\centering
\subfigure []{\label{fig:9.a}\includegraphics[height=0.25\textwidth]{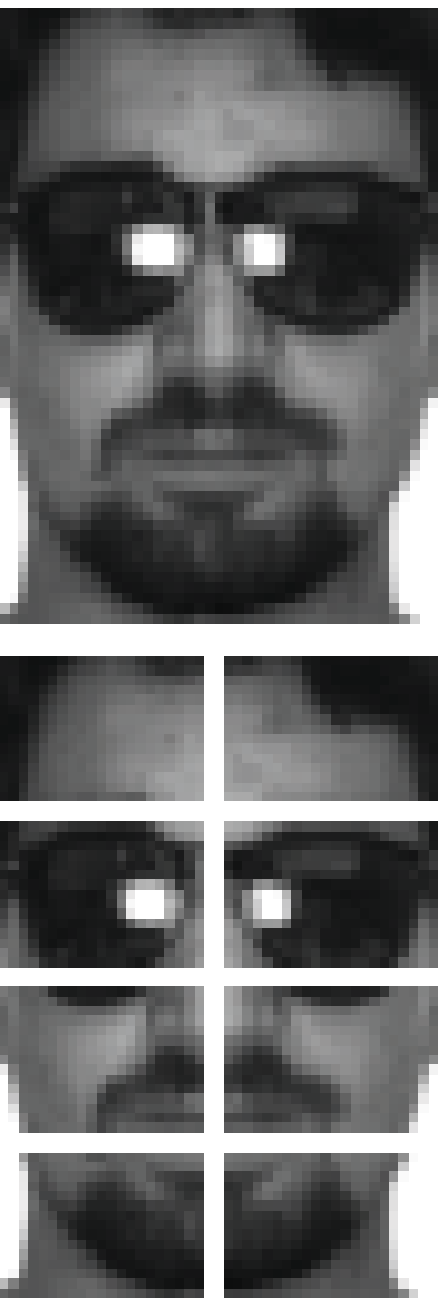}}\hspace{0.1cm}
\subfigure []{\label{fig:9.b}\includegraphics[width=0.36\textwidth]{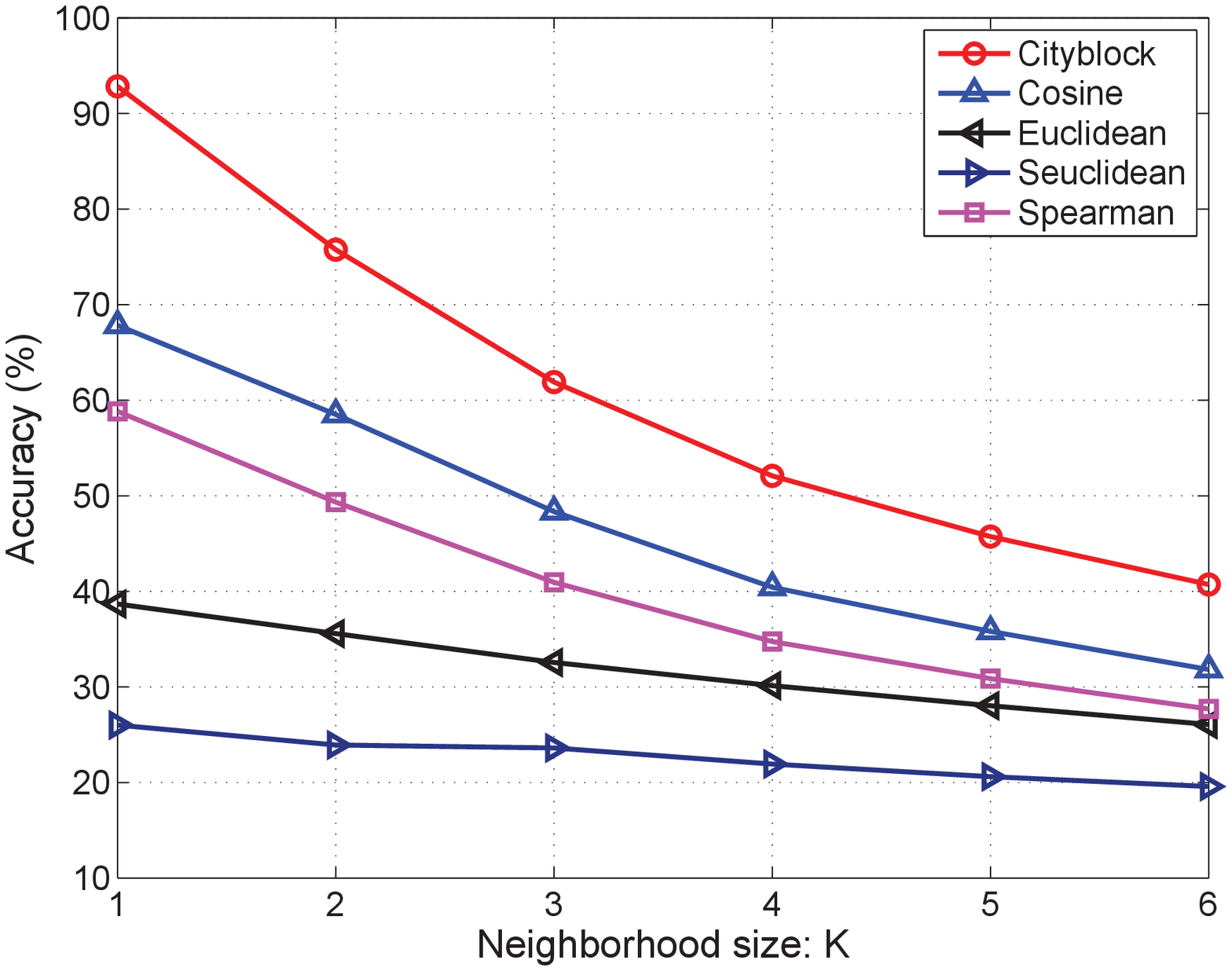}}\hspace{0.1cm}
\subfigure []{\label{fig:9.c}\includegraphics[height=0.25\textwidth]{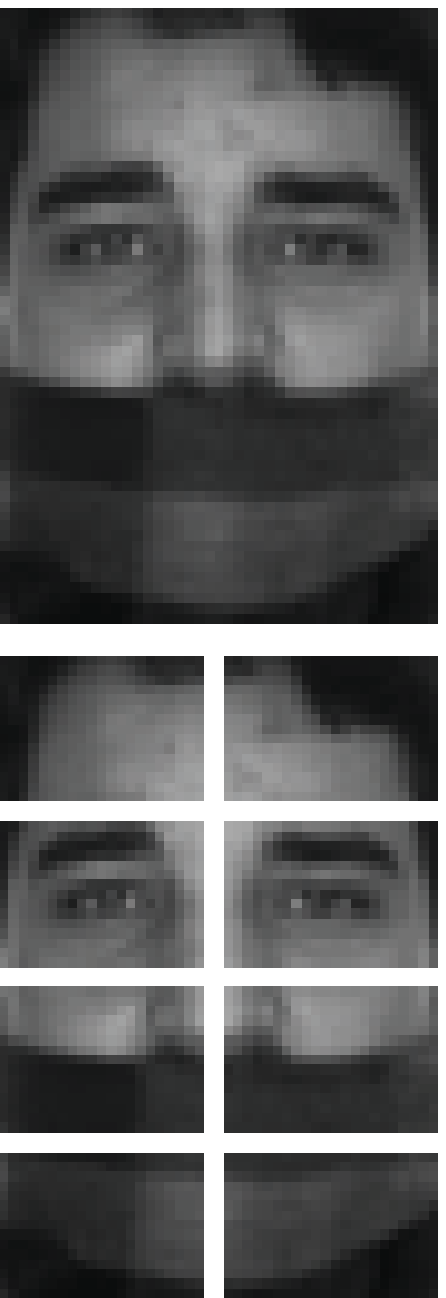}}\hspace{0.1cm}
\subfigure []{\label{fig:9.d}\includegraphics[width=0.36\textwidth]{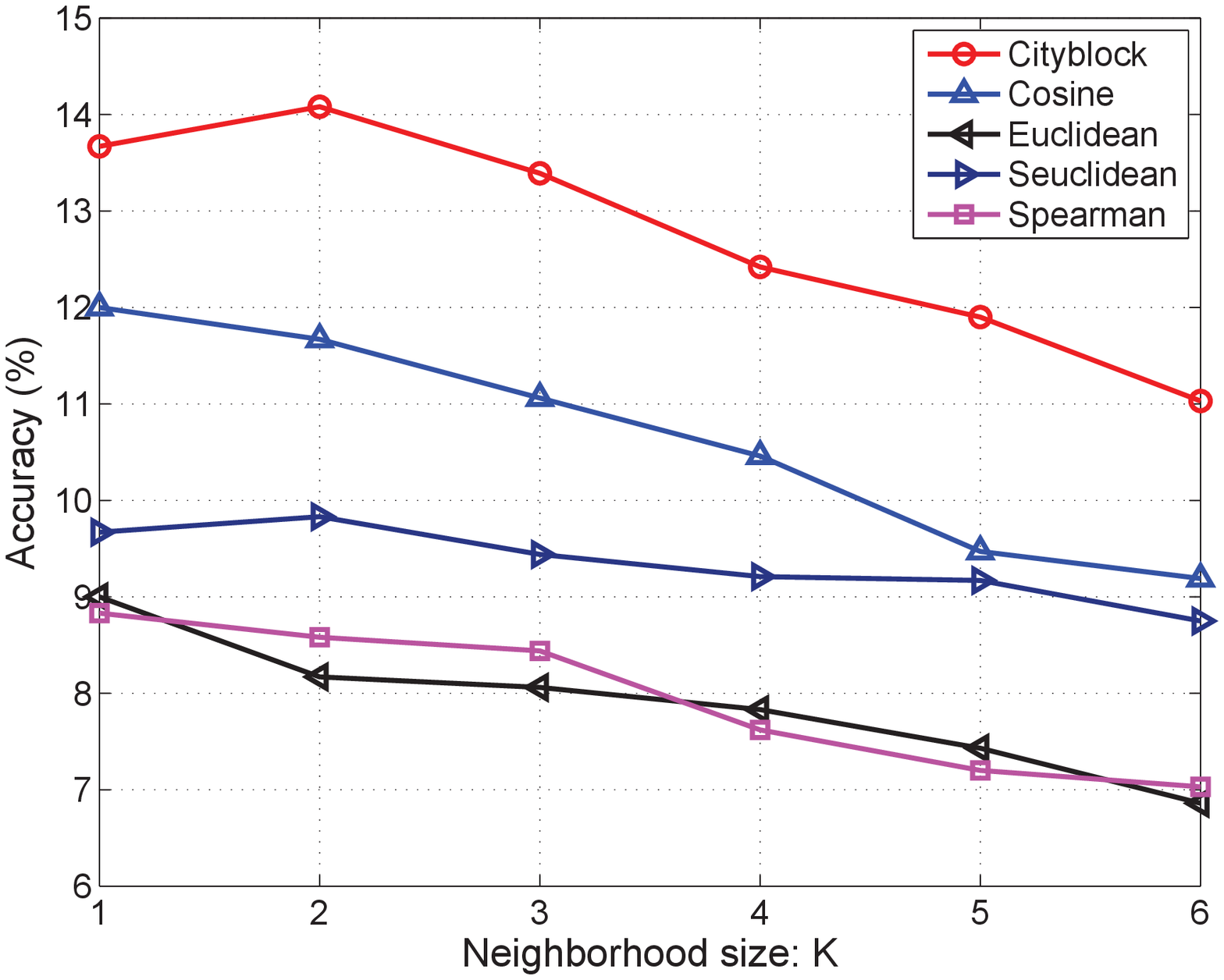}}\hspace{0.1cm}
\begin{scriptsize}
\subtable []{
\label{fig:9.e}
\begin{tabular}{|c||c|c|c|c|c |c|c|}
\hline
\bfseries Disguise & \multicolumn{3}{c|}{\bfseries sunglasses} & \multicolumn{3}{c|}{\bfseries scarves}\\
\hline
\bfseries Feature & \multicolumn{2}{c|}{\bfseries Holistic} & \multicolumn{1}{c|}{\bfseries Partitioned}
& \multicolumn{2}{c|}{\bfseries Holistic} & \multicolumn{1}{c|}{\bfseries Partitioned}\\
\hline
\bfseries Dim &   \bfseries 300 & \bfseries 2580 & \bfseries 2580
&  \bfseries 300 & \bfseries 2580 & \bfseries 2580\\
\hline\hline
SVM~\cite{Fan2008}    &    47.83\% &  48.67\% &  40.17\%
                      &    13.50\% &  13.83\% &  41.67\% \\
LPP~\cite{He2003-Locality} & 14.83\% & 18.50\% & 88.83\%
                           & 20.33\% & 24.17\% & 82.00\%\\
SRC~\cite{Wright2009-Robust}    &    57.00\% & -  &  93.00\%
                                &  \bfseries   \bfseries 69.83\% & -  &  91.83\%  \\
CESR~\cite{He2011-Maximum} & 21.17\% & \bfseries95.50\% & \bfseries97.50\%
                           & 31.50\% & 16.67\% & 91.33\%\\
LRC~\cite{Shi2011-recognition,Naseem2010-Linear}  &   52.83\% &  49.17\% &  88.17\%
                                                  &   68.50\% &  57.50\% &  91.83\%\\
CRC-RLS~\cite{Zhang2011-Sparse}   &    53.00\% &  71.50\% &  88.33\%
                                  &    68.50\% &  89.17\% &  92.17\%\\
\hline
LCCR + Cityblock  & \bfseries 93.00\% & 93.50\% &  91.17\%
                   & 68.67\% & 89.17\% &  92.50\%\\
LCCR + Seuclidean   &  57.83\% &  74.50\% &  91.50\%
                    &  68.50\% &  89.17\% &  92.17\%\\
LCCR + Euclidean   &  66.00\% &  79.00\% &  90.83\%
                 &  68.67\% &  89.17\% &  92.50\%\\
LCCR + Cosine &   76.83\% &  85.33\% &  93.83\%
               &   68.83\% &  \bfseries 89.50\% &  \bfseries 93.83\%\\
LCCR + Spearman &  70.67\% &  81.17\% & 95.83\%
                 &  68.67\% &  89.33\% & \bfseries 93.83\%\\
\hline
\end{tabular}}
\end{scriptsize}
\caption{Recognition on AR faces with real possible occlusions.
(a) The top row is a facial image occluded by sunglass, whose
partitioned blocks are shown as below. (b) The accuracy of
\emph{K}-NN searching using Cityblock distance, Cosine distance,
Euclidean distance, Seuclidean distance and Spearman distance on
the AR images with sunglasses (2580D). (c) Similar to (a), the top
row is a face occluded by scarf, and its partitions below. (d) The
precision of \emph{K}-NN searching using Cityblock, Cosine,
Euclidean, Seuclidean, Spearman as distance metrics on the AR
images with scarves (2580D). (e) The recognition rates of
competing methods across different experimental configurations.}
\label{fig:9}
\end{figure*}

On the other hand, it is easy to find that LRC, CRC-RLS
and LCCRs are more robust than SRC and SVM, which implies that the
$\ell^1$-regularization term cannot yield better robustness than
the $\ell^2$-regularization term, at least for the Eigenface
space. Moreover, the models achieve better results in higher
dimensional space, even though the difference of classification
accuracy between higher dimensional space and lower ones is less than
$1\%$ except CESR has an obvious improvement.

\subsection{Face Recognition with Real Occlusions}
\label{sec:4.7}

In this sub-section, we examine the robustness to real possible
occlusions of the investigated approaches over the AR data set. We use 1400 clean images for
training, 600 faces wearing by sunglasses (occluded ratio is about
20\%) and 600 face wearing by scarves (occluded ratio is about
40\%) for testing, separately. In~\cite{Wright2009-Robust}, Wright et al. only used a third of disguised images for this test, i.e., 200 images for each kind of disguises. In addition, we also investigate the role of \emph{K}-NN searching in LCCR.

We examine two widely-used feature schemes, namely, the holistic feature with 300D and 2580D, as well as the partitioned
feature based on the cropped data. The partitioned feature scheme firstly partitions an image into multiple blocks (8
blocks as did in~\cite{Wright2009-Robust,Zhang2011-Sparse,Yang2012-Relaxed}, see \figurename~\ref{fig:9.a} and~\ref{fig:9.c}), then conducts
classification on each block independently, and after that, aggregates the results by voting.

\figurename~\ref{fig:9.e} reports the recognition rates of all the tested methods. For the images occluded by sunglasses, LCCR with
Cityblock distance and CESR achieve remarkable results with the holistic feature scheme, their recognition accuracy are nearly double that of
the other methods. This considerable performance margin contributes to the accuracy of $K$-NN searching based on Cityblock distance (see \figurename~\ref{fig:9.b}).

For the images occluded by scarves, LCCR achieves the highest recognition rate over the full dimensional space, and the second highest rates using Eigenface. However, the difference in rates between LCCR and other non-iterative algorithms (LRC, CRC-RLS) is very small due to the poor accuracy of \emph{K}-NN searching as shown in \figurename~\ref{fig:9.d}. Furthermore, the partitioned feature scheme produces higher recognition rates than the holistic one for all
competing methods, which is consistent with previous report~\cite{Wright2009-Robust}.

From the above experiments, it is easy to conclude that the preservation of locality is helpful to coding scheme, especially
when the real structures of data cannot be found by traditional coding scheme. Moreover, the performance ranking of LCCR with five
distance metrics is same with that of \emph{K}-NN searching with the used metrics.

\subsection{Face Recognition with Corruption}
\label{sec:4.6}

\begin{figure}[t]
\centering
\includegraphics[width=0.41\textwidth]{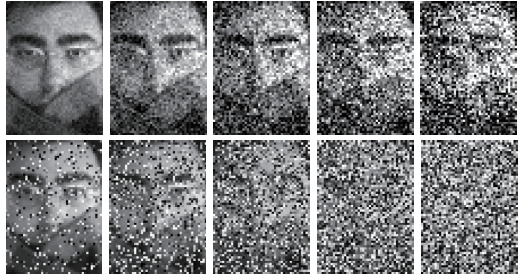}
\caption{Testing images from AR database with additive noise and non-additive noise. Top row: 10\%, 30\%, 50\%, 70\%, 90\% white noises are added into test image; Bottom row: the case of random pixel corruption with 10\%-90\% percentages, respectively.}
\label{fig:8}
\end{figure}

We test the robustness of LCCR against two kinds of corruption
using the AR data set containing 2600 images of 100 individuals.
For each subject, we use 13 images for training (7 clean images, 3
images with sunglasses, and 3 images with scarves), and the
remaining 13 images for testing. Different from~\cite{Wright2009-Robust}
which tested the robustness to corruption using the Extended Yale
B database, our case is more challenging for the following
reasons. Firstly, AR images contain real possible occlusions, i.e.,
sunglasses and scarves, while Extended Yale B is a set of clean
images without disguises. Secondly, AR includes more facial
variations (13 versus 9), more subjects (100 versus 38), and a
smaller samples for each subject (26 images per subject versus 64
images per subject). Thirdly, we investigated two kinds of
corruption, white noise (additive noise) and random pixel
corruption (non-additive noise) which are two commonly assumed in
face recognition problem~\cite{Wright2009-Robust,Shi2011-recognition,Naseem2010-Linear}. For the white
noise case (the top row of \figurename~\ref{fig:8}), we add random
noise from normal distribution to each testing image $\mathbf{x}$,
that is, $\mathbf{\tilde{x}} = \mathbf{x}+\alpha \mathbf{n}$, and
restrict $\mathbf{\tilde{x}}\in[0\ 255]$, where $\alpha$ is the
corruption ratio from $10\%$ to $90\%$ with an interval of
$20\%$, and $\mathbf{n}$ is the noise following a standard normal
distribution. For the random pixel corruption case (the bottom row
in \figurename~\ref{fig:8}), we replace the value of a percentage
of pixels randomly chosen from each test image with the values
following a uniform distribution over $[0 \ p_{max}]$, where
$p_{max}$ is the largest pixel value of current image.

To improve the anti-noise ability of SRC~\cite{Wright2009-Robust}, Wright
et al. generate a new dictionary $[\mathbf{D}\ \mathbf{I}]$ by
concatenating an identity matrix $\mathbf{I}$ with the original
dictionary $\mathbf{D}$, where the dimensionality of $\mathbf{I}$
equals to that of data. The use of $\mathbf{I}$ has been verified
to be effective in improving the robustness of $\ell^1$-norm based
models~\cite{Qiao2010Sparsity, Wright2009-Robust} at the cost of time-consuming. Therefore, it is a tradeoff between
robustness and efficiency for the algorithms. Will
the strategy still work for $\ell^2$-minimization based models?
In this sub-section, we fill this gap by comparing the results by
coding over these two dictionary.

\begin{table*}[t]
\caption{The Robustness of Different Methods over AR Database with 300D (Coding over $\mathbf{D}$).}
\label{tab:5} \centering
\begin{scriptsize}
\begin{tabular}{|c||c|c|c|c|c| c|c|c|c|c|}
\hline
\bfseries Corruptions & \multicolumn{5}{c|}{\bfseries White Gaussian Noise} & \multicolumn{5}{c|}{\bfseries Random Pixel Corruption}\\
\hline
\bfseries Corrupted ratio &  \bfseries 10\% & \bfseries 30\% & \bfseries 50\% & \bfseries 70\% & \bfseries 90\% &  \bfseries 10\% & \bfseries 30\% & \bfseries 50\% & \bfseries 70\% & \bfseries 90\%\\
\hline\hline
SVM~\cite{Fan2008} & 91.77\% & 91.38\% & 90.23\% & 88.62\% & 82.69\% & 91.54\% & 81.92\% & 45.46\% & 8.23\% & 2.00\%\\
LPP+1NN~\cite{He2003-Locality} & 29.31\% & 8.46\% & 4.08\% & 2.38\% & 2.62\% & 5.69\% & 2.62\% & 2.00\% & 1.46\% & 1.17\%\\
SRC~\cite{Wright2009-Robust} & 92.62\% & 91.23\% & 86.54\% & 78.31\% & 62.62\% & 89.62\% & 72.31\% & 38.85\% & 8.23\% & 2.00\%\\
CESR~\cite{He2011-Maximum} & 89.69\% & 87.85\% & 85.38\% & 80.85\% & 73.00\% & 87.38\% & 76.31\% & 43.23\% & 12.38\% & 1.46\%\\
LRC~\cite{Shi2011-recognition,Naseem2010-Linear} & 93.39\% & 92.39\% & 88.85\% & 81.85\% & 67.62\% & 91.77\% & 77.00\% & 45.77\% & 13.62\% & 2.54\%\\
CRC-RLS~\cite{Zhang2011-Sparse} & 94.77\% & 94.39\% & 92.85\% & 90.92\% & 87.31\% & 84.08\% & 88.69\% & 65.46\% & 20.77\% & 2.92\%\\
\hline
LCCR + Cityblock & \bfseries97.00\% & \bfseries96.00\% & 94.54\% & 92.31\% & 89.08\% & \bfseries96.54\% & 92.31\% & 79.69\% & 37.08\% & 5.23\%\\
LCCR + Seuclidean & 96.31\% & 95.85\% & 94.46\% & 92.39\% & 88.54\% & 95.69\% & 90.08\% & 65.85\% & 20.77\% & 3.00\%\\
LCCR + Euclidean & 95.77\% & 95.23\% & 94.23\% & 92.31\% & 88.31\% & 95.39\% & 90.39\% & 67.23\% & 20.92\% & 3.23\%\\
LCCR + Cosine & 95.62\% & 95.31\% & 93.92\% & 92.15\% & 88.69\% & 94.85\% & 89.46\% & 65.62\% & 20.85\% & 3.69\%\\
LCCR + Spearman & 96.15\% & 95.39\% & \bfseries94.69\% & \bfseries93.08\% & \bfseries89.77\% & 95.54\% & \bfseries92.54\% & \bfseries83.00\% & \bfseries59.31\% & \bfseries13.69\%\\
\hline
\end{tabular}
\end{scriptsize}
\end{table*}

\begin{table*}[t]
\caption{The Robustness of Different Methods over AR Database with 2580D (Coding over $\mathbf{D}$).}
\label{tab:6} \centering
\begin{scriptsize}
\begin{tabular}{|c||c|c|c|c|c| c|c|c|c|c|}
\hline
\bfseries Corruptions & \multicolumn{5}{c|}{\bfseries White Gaussian Noise} & \multicolumn{5}{c|}{\bfseries Random Pixel Corruption}\\
\hline
\bfseries Corrupted ratio &  \bfseries 10\% & \bfseries 30\% & \bfseries 50\% & \bfseries 70\% & \bfseries 90\% &  \bfseries 10\% & \bfseries 30\% & \bfseries 50\% & \bfseries 70\% & \bfseries 90\%\\
\hline\hline
SVM~\cite{Fan2008} & 91.92\% & 91.31\% & 89.85\% & 88.62\% & 82.92\% & 91.69\% & 81.46\% & 44.77\% & 7.69\% & 1.92\%\\
LPP+1NN~\cite{He2003-Locality} & 37.69\% & 10.31\% & 4.54\% & 3.00\% & 2.54\% & 6.92\% & 2.54\% & 2.15\% & 1.46\% & 1.47\%\\
CESR~\cite{Wright2009-Robust} & 90.85\% & 86.69\% & 84.38\% & 78.85\% & 70.08\% & 91.00\% & 90.77\% & \bfseries90.54\% & \bfseries66.08\% & 13.08\%\\
LRC~\cite{Shi2011-recognition,Naseem2010-Linear} & 78.69\% & 33.77\% & 4.62\% & 4.62\% & 2.69\% & 21.77\% & 3.77\% & 2.39\% & 0.92\% & 1.15\%\\
CRC-RLS~\cite{Zhang2011-Sparse} & 94.85\% & 94.77\% & 93.23\% & 90.85\% & 88.39\% & 94.15\% & 89.08\% & 67.08\% & 22.69\% & 2.62\%\\
\hline
LCCR + Cityblock & \bfseries97.54\% & 96.08\% & 95.08\% & 93.15\% & \bfseries90.54\% & \bfseries96.85\% & \bfseries93.23\% & 78.77\% & 29.77\% & 4.54\%\\
LCCR + Seuclidean & 96.92\% & \bfseries96.23\% & \bfseries95.39\% & 92.92\% & 89.00\% & 96.00\% & 90.77\% & 67.31\% & 22.69\% & 3.00\%\\
LCCR + Euclidean & 96.08\% & 95.62\% & 94.85\% & 92.23\% & 88.92\% & 95.62\% & 91.15\% & 68.31\% & 22.92\% & 3.00\%\\
LCCR + Cosine & 96.08\% & 95.46\% & 94.39\% & 92.54\% & 89.46\% & 95.31\% & 90.77\% & 67.15\% & 23.23\% & 3.62\%\\
LCCR + Spearman & 96.54\% & 95.23\% & 94.69\% & \bfseries93.31\% & 90.39\% & 95.85\% & 92.92\% & 83.31\% & 60.69\% & \bfseries13.85\%\\
\hline
\end{tabular}
\end{scriptsize}
\end{table*}

\begin{table*}[t]
\caption{The Robustness of Different Methods over AR Database with 300D (Coding over [$\mathbf{D}$ $\mathbf{E}$]).}
\label{tab:7} \centering
\begin{scriptsize}
\begin{tabular}{|c||c|c|c|c|c| c|c|c|c|c|}
\hline
\bfseries Corruptions & \multicolumn{5}{c|}{\bfseries White Gaussian Noise} & \multicolumn{5}{c|}{\bfseries Random Pixel Corruption}\\
\hline
\bfseries Corrupted ratio &  \bfseries 10\% & \bfseries 30\% & \bfseries 50\% & \bfseries 70\% & \bfseries 90\% &  \bfseries 10\% & \bfseries 30\% & \bfseries 50\% & \bfseries 70\% & \bfseries 90\%\\
\hline\hline
SRC~\cite{Wright2009-Robust} & 92.62\% & 91.08\% & 90.46\% & 87.92\% & 84.08\% & 91.62\% & 83.38\% & 56.31\% & 14.92\% & 2.08\%\\
CESR~\cite{He2011-Maximum} & 85.69\% & 82.46\% & 80.31\% & 73.69\% & 63.15\% & 83.38\% & 69.00\% & 34.69\% & 10.62\% & 2.92\%\\
LRC~\cite{Shi2011-recognition,Naseem2010-Linear} & 92.62\% & 92.15\% & 91.00\% & 88.85\% & 86.31\% & 91.15\% & 84.46\% & 49.92\% & 8.31\% & 1.85\%\\
CRC-RLS~\cite{Zhang2011-Sparse} & 92.62\% & 92.15\% & 91.00\% & 88.85\% & 86.31\% & 91.15\% & 84.46\% & 49.92\% & 8.31\% & 1.85\%\\
\hline
LCCR + Cityblock & \bfseries93.08\% & 92.39\% & 91.69\% & 89.62\% & 86.62\% & 92.62\% & 87.54\% & 69.46\% & 34.00\% & \bfseries4.92\%\\
LCCR + Seuclidean & 92.85\% & 92.39\% & 91.69\% & 89.62\% & 86.46\% & 91.85\% & 85.23\% & 50.00\% & 8.31\% & 2.46\%\\
LCCR + Euclidean & 92.85\% & 92.39\% & 91.69\% & 89.92\% & 86.62\% & 91.92\% & 85.15\% & 51.23\% & 10.23\% & 2.54\%\\
LCCR + Cosine & 92.69\% & 92.69\% & 91.92\% & 89.92\% & 87.00\% & 92.15\% & 84.92\% & 49.92\% & 8.38\% & 2.62\%\\
LCCR + Spearman & 92.85\% & \bfseries92.85\% & \bfseries92.15\% & \bfseries90.15\% & \bfseries87.54\% & \bfseries92.69\% & \bfseries87.85\% & \bfseries75.92\% & \bfseries56.54\% & 1.38\%\\
\hline
\end{tabular}
\end{scriptsize}
\end{table*}

\begin{table*}[t]
\caption{The Robustness of Different Methods over AR Database with 2580D (Coding over [$\mathbf{D}$ $\mathbf{E}$]).}
\label{tab:8} \centering
\begin{scriptsize}
\begin{tabular}{|c||c|c|c|c|c| c|c|c|c|c|}
\hline
\bfseries Corruptions & \multicolumn{5}{c|}{\bfseries White Gaussian Noise} & \multicolumn{5}{c|}{\bfseries Random Pixel Corruption}\\
\hline
\bfseries Corrupted ratio &  \bfseries 10\% & \bfseries 30\% & \bfseries 50\% & \bfseries 70\% & \bfseries 90\% &  \bfseries 10\% & \bfseries 30\% & \bfseries 50\% & \bfseries 70\% & \bfseries 90\%\\
\hline\hline
CESR~\cite{He2011-Maximum} & 91.85\% & 87.15\% & 82.08\% & 71.85\% & 57.62\% & 91.38\% & \bfseries 91.15\% & \bfseries 90.23\% & \bfseries 76.85\% & 8.08\%\\
LRC~\cite{Shi2011-recognition,Naseem2010-Linear} & 93.00\% & 92.46\% & 91.69\% & 90.15\% & 87.15\% & 92.15\% & 84.92\% & 51.15\% & 8.54\% & 1.77\%\\
CRC-RLS~\cite{Zhang2011-Sparse} & 93.00\% & 92.46\% & 91.69\% & 89.23\% & 87.15\% & 92.15\% & 84.92\% & 51.15\% & 8.54\% & 1.77\%\\
\hline
LCCR + Cityblock & \bfseries 93.39\% & 93.00\% & 92.15\% & 90.15\% & 87.39\% & 93.08\% & 88.23\% & 70.15\% & 34.08\% & 5.08\%\\
LCCR + Seuclidean & 93.31\% & 93.00\% & 92.15\% & 90.15\% & 87.31\% & 92.69\% & 85.85\% & 51.39\% & 8.62\% & 2.46\%\\
LCCR + Euclidean & 93.31\% & 92.77\% & 91.92\% & 90.31\% & 87.62\% & 92.62\% & 86.00\% & 52.46\% & 10.69\% & 2.54\%\\
LCCR + Cosine & 93.15\% & \bfseries 93.23\% & 92.23\% & 90.39\% & 87.77\% & 92.69\% & 85.62\% & 51.59\% & 8.62\% & 2.77\%\\
LCCR + Spearman & \bfseries 93.39\% & 93.15\% & \bfseries 92.62\% & \bfseries 90.54\% & \bfseries 88.31\% & \bfseries 93.31\% & 88.54\% & 77.00\% & 56.54\% & \bfseries 13.62\%\\
\hline
\end{tabular}
\end{scriptsize}
\end{table*}

\tablename~\ref{tab:5} through \tablename~\ref{tab:8} are the recognition rates of the tested methods across feature space
(Eigenface with 300D) and full dimensional space (2580D). We didn't reported the results of SVM and LPP with the strategy of expanding dictionary since the methods are not belong to the facility of linear coding scheme. Moreover, SRC requires the dictionary is an over-completed matrix such that it could not run in the full dimensional cases. Based on the results, we have the following conclusions:

Firstly, the proposed LCCRs are much superior to SVM, LPP, SRC, CESR, LRC and CRC-RLS. For
example, in the worst case (the white gaussian noise corruption ratio is $90\%$, the best result of
LCCR is about $90.54\%$ (\tablename~\ref{tab:6}), compared to $82.92\%$ of SVM (\tablename~\ref{tab:6}), $2.62\%$ of LPP (\tablename~\ref{tab:5}) , $84.08\%$ of SRC (\tablename~\ref{tab:7}), $73\%$ of CESR (\tablename~\ref{tab:5}), $87.15\%$ of LRC (\tablename~\ref{tab:8}), and $88.39\%$ of CRC-RLS
(\tablename~\ref{tab:6}). In the case of random pixel corruption, one can see when the corruption ratio reaches $70\%$, all methods fail to perform recognition except LCCR in the two data spaces and CESR in the full dimensional space.

Secondly, all investigated algorithms perform worse with increased corruption ratio and achieve better results in white noise
corruption (additive noise) than random pixel corruption (non-additive noise). Moreover, the improvement of CESR is obvious when the original data is used to test. As discussed in the above, the improvement is at the cost of computational efficiency. For the other methods, they perform slightly better
(less than $1\%$) in the full-dimensional space except LRC.

Thirdly, the results show that coding over $[\mathbf{D}\ \mathbf{I}]$ is helpful in improving the robustness of SRC and
LRC, but it has negative impact on the recognition accuracy of CESR, CRC-RLS and LCCR. For example, when white noise ratio
rises to $90\%$ for the Eigenface (\tablename~\ref{tab:5}, expanding $\mathbf{D}$ leads to the variation of the recognition rate from $62.62\%$ to $84.08\%$ for SRC, from $73.00\%$ to $63.15\%$ for CESR, from $67.62\%$ to $86.31\%$ for LRC, from $87.31\%$ to
$86.31\%$ for CRC-RLS, and from $89.77\%$ to $87.54\%$ for LCCR with Spearman distance. The conclusion has not been reported in the
previous works.

\section{Conclusions and Discussions}
\label{sec:5}

It is interesting and important to improve the discrimination and robustness of data representation. The traditional coding
algorithm gets the representation by encoding each datum as a linear combination of a set of training samples, which mainly depicts the global
structure of data. However, it will be failed when the data are grossly corrupted. Locality (Local consistency) preservation, which keeps the geometric structure of manifold for dimension reduction, has shown the effectiveness in revealing the real structure of data. In this paper, we proposed a novel objective function to get an effective and robust representation by enforcing the similar inputs produce similar codes, and the function possesses analytic
solution.

The experimental studies showed that the introduction of locality makes LCCR more accurate and robust to various occlusions and corruptions. We investigated the performance of LCCR with five basic distance metrics (for locality). The results imply that if better \emph{K}-NN searching methods or more sophisticated
distance metrics are adopted, LCCR might achieve a higher recognition rate. Moreover, the performance comparisons over two different dictionaries show that it is unnecessary to expand the dictionary $\mathbf{D}$ with $\mathbf{I}$ for $\ell^2$-norm based coding algorithms.

Each approach has its own advantages and disadvantages. Parameter determination maybe is the biggest problem of LCCR which requires three user-specified parameters. In the future works, it is possible to explore the relationship between locality parameter $k$ and the intrinsic dimensionality of sub-manifold. Moreover, the work has focused on the representation learning, however, dictionary learning is also important and interesting in this area. Therefore, an possible way to extend this work is exploring how to reflect local consistency in the formation process of dictionary.

\bibliographystyle{elsarticle-num}
\bibliography{LCCR}

\begin{thebibliography}{10}
\expandafter\ifx\csname url\endcsname\relax
  \def\url#1{\texttt{#1}}\fi
\expandafter\ifx\csname urlprefix\endcsname\relax\def\urlprefix{URL }\fi
\expandafter\ifx\csname href\endcsname\relax
  \def\href#1#2{#2} \def\path#1{#1}\fi

\bibitem{Candes2005-Decoding}
E.~J. Candes, T.~Tao, Decoding by linear programming, IEEE Transactions on
  Information Theory 51~(12) (2005) 4203--4215.

\bibitem{Donoho2006-large}
D.~L. Donoho, For most large underdetermined systems of linear equations the
  minimal $\ell^{1}$-norm solution is also the sparsest solution,
  Communications on Pure and Applied Mathematics 59~(6) (2006) 797--829.

\bibitem{Chen2001-Atomic}
S.~S.~B. Chen, D.~L. Donoho, M.~A. Saunders, Atomic decomposition by basis
  pursuit, SIAM Review 43~(1) (2001) 129--159.

\bibitem{Efron2004-Least}
B.~Efron, T.~Hastie, I.~Johnstone, R.~Tibshirani, Least angle regression,
  Annals of Statistics 32~(2) (2004) 407--451.

\bibitem{Yang2010-l1-minimization}
A.~Yang, A.~Ganesh, S.~Sastry, Y.~Ma, Fast $\ell^{1}$-minimization algorithms
  and an application in robust face recognition: a review, in: Proc. of
  International Conference on Image Processing, 2010, pp. 1849--1852.

\bibitem{Peng2012}
P.~Xi, L.~Zhang, Z.~Yi., Constructing l2-graph for subspace learning and
  segmentation, ArXiv e-prints\href {http://arxiv.org/abs/1209.0841}
  {\path{arXiv:1209.0841}}.

\bibitem{Cheng2010-Learning}
B.~Cheng, J.~Yang, S.~Yan, Y.~Fu, T.~Huang, Learning with $\ell^{1}$-graph for
  image analysis, IEEE Transactions on Image Processing 19~(4) (2010) 858--866.

\bibitem{Elhamifar2012-Sparse}
E.~Elhamifar, R.~Vidal, Sparse subspace clustering: Algorithm, theory, and
  applications, To appear in IEEE Transactions on Pattern Analysis and Machine
  Intelligence.

\bibitem{Wang2011-Image}
C.~Wang, X.~He, J.~Bu, Z.~Chen, C.~Chen, Z.~Guan, Image representation using
  laplacian regularized nonnegative tensor factorization, Pattern Recognition
  44~(10) (2011) 2516--2526.

\bibitem{Wright2009-Robust}
J.~Wright, A.~Y. Yang, A.~Ganesh, S.~S. Sastry, Y.~Ma, Robust face recognition
  via sparse representation, IEEE Transactions on Pattern Analysis and Machine
  Intelligence 31~(2) (2009) 210--227.

\bibitem{Zhang2012Joint}
H.~Zhang, N.~M. Nasrabadi, Y.~Zhang, T.~S. Huang, Joint dynamic sparse
  representation for multi-view face recognition, Pattern Recognition 45~(4)
  (2012) 1290--1298.
\newblock \href {http://dx.doi.org/10.1016/j.patcog.2011.09.009}
  {\path{doi:10.1016/j.patcog.2011.09.009}}.

\bibitem{Zhang2013-Simultaneous}
H.~Zhang, Y.~Zhang, T.~S. Huang, Simultaneous discriminative projection and
  dictionary learning for sparse representation based classification, Pattern
  Recognition 46~(1) (2013) 346--354.

\bibitem{He2011-Maximum}
R.~He, W.-S. Zheng, B.-G. Hu, Maximum correntropy criterion for robust face
  recognition, IEEE Transactions on Pattern Analysis and Machine Intelligence
  33~(8) (2011) 1561--1576.

\bibitem{Li1999-recognition}
S.~Z. Li, J.~Lu, Face recognition using the nearest feature line method, IEEE
  Transactions on Neural Networks 10~(2) (1999) 439--443.

\bibitem{Yang2012-Beyond}
J.~Yang, L.~Zhang, Y.~Xu, J.~Yang, Beyond sparsity: The role of l1-optimizer in
  pattern classification, Pattern Recognition 45~(3) (2012) 1104--1118.

\bibitem{Rigamonti2011-sparse}
R.~Rigamonti, M.~A. Brown, V.~Lepetit, Are sparse representations really
  relevant for image classification?, in: Proc. of IEEE International
  Conference on Computer Vision and Pattern Recognition, 2011, pp. 1545--1552.

\bibitem{Shi2011-recognition}
Q.~Shi, A.~Eriksson, A.~van~den Hengel, C.~Shen, Is face recognition really a
  compressive sensing problem?, in: Proc. of IEEE Conference on Computer Vision
  and Pattern Recognition.

\bibitem{Zhang2011-Sparse}
L.~Zhang, M.~Yang, X.~Feng, Sparse representation or collaborative
  representation: Which helps face recognition?, in: Proc. of IEEE
  International Conference on Computer Vision.

\bibitem{Naseem2010-Linear}
I.~Naseem, R.~Togneri, M.~Bennamoun, Linear regression for face recognition,
  IEEE Transactions on Pattern Analysis and Machine Intelligence 32~(11) (2010)
  2106--2112.

\bibitem{He2005-Neighborhood}
X.~He, D.~Cai, S.~Yan, H.~Zhang, Neighborhood preserving embedding, in: Proc.
  of IEEE International Conference on Computer Vision.

\bibitem{Belkin2006-Manifold}
M.~Belkin, P.~Niyogi, V.~Sindhwani, Manifold regularization: A geometric
  framework for learning from labeled and unlabeled examples, The Journal of
  Machine Learning Research 7 (2006) 2399--2434.

\bibitem{Yan2007-Graph}
S.~C. Yan, D.~Xu, B.~Y. Zhang, H.~J. Zhang, Q.~Yang, S.~Lin, Graph embedding
  and extensions: A general framework for dimensionality reduction, IEEE
  Transactions on Pattern Analysis and Machine Intelligence 29~(1) (2007)
  40--51.

\bibitem{Richard2006-Random}
R.~G. Baraniuk, M.~B. Wakin, Random projections of smooth manifolds,
  Foundations of Computational mathematics 9~(1) (2009) 51--77.

\bibitem{Majumdar2010-Robust}
A.~Majumdar, R.~K. Ward, Robust classifiers for data reduced via random
  projections, IEEE Transactions on Systems, Man, and Cybernetics, Part B:
  Cybernetics 40~(5) (2010) 1359--1371.
\newblock \href {http://dx.doi.org/10.1109/TSMCB.2009.2038493}
  {\path{doi:10.1109/TSMCB.2009.2038493}}.

\bibitem{Wang2010-Locality}
J.~Wang, J.~Yang, K.~Yu, F.~Lv, T.~Huang, Y.~Gong, Locality-constrained linear
  coding for image classification, in: Proc. of IEEE International Conference
  on Computer Vision and Pattern Recognition, 2010, pp. 3360--3367.

\bibitem{Roweis2000}
S.~T. Roweis, L.~K. Saul, Nonlinear dimensionality reduction by locally linear
  embedding, Science 290~(5500) (2000) 2323--2326.

\bibitem{Chao2011-Locality}
Y.~Chao, Y.~Yeh, Y.~Chen, Y.~Lee, Y.~Wang, Locality-constrained group sparse
  representation for robust face recognition, in: Proc. of IEEE International
  Conference on Image Processing, 2011, pp. 761--764.

\bibitem{Yang2012-Relaxed}
M.~Yang, L.~Zhang, D.~Zhang, S.~Wang, Relaxed collaborative representation for
  pattern classification, in: Proc. of IEEE Conference on Computer Vision and
  Pattern Recognition, 2012, pp. 2224--2231.

\bibitem{Ohki2005}
K.~Ohki, S.~Chung, Y.~H. Ch'ng, P.~Kara, R.~C. Reid, Functional imaging with
  cellular resolution reveals precise micro-architecture in visual cortex,
  Nature 433~(7026) (2005) 597--603.

\bibitem{He2003-Locality}
X.~He, S.~Yan, Y.~Hu, P.~Niyogi, H.~Zhang, Face recognition using
  laplacianfaces, IEEE Transactions on Pattern Analysis and Machine
  Intelligence 27~(3) (2005) 328--340.

\bibitem{Osborne2000}
M.~R. Osborne, B.~Presnell, B.~A. Turlach, A new approach to variable selection
  in least squares problems, IMA Journal of Numerical Analysis 20~(3) (2000)
  389--403.

\bibitem{Peng2013}
X.~Peng, L.~Zhang, Z.~Yi, Scalable sparse subspace clustering, in: Proc. of
  IEEE Conference on Computer Vision and Pattern Recognition, 2013.

\bibitem{Martinez1998}
A.~Martinez, R.~Benavente, The ar face database (1998).

\bibitem{Samaria1994}
F.~Samaria, A.~Harter, Parameterisation of a stochastic model for human face
  identification, in: Proc. of the IEEE Workshop on Applications of Computer
  Vision (WACV), 1994, pp. 138--142.

\bibitem{Georghiades2001}
A.~Georghiades, P.~Belhumeur, D.~Kriegman, From few to many: illumination cone
  models for face recognition under variable lighting and pose, IEEE
  Transactions on Pattern Analysis and Machine Intelligence 23~(6) (2001)
  643--660.

\bibitem{Gross2010}
R.~Gross, I.~Matthews, J.~Cohn, T.~Kanade, S.~Baker, Multi-pie, Image and
  Vision Computing 28~(5) (2010) 807--813.

\bibitem{Fan2008}
R.-E. Fan, K.-W. Chang, C.-J. Hsieh, X.-R. Wang, C.-J. Lin, Liblinear: A
  library for large linear classification, Journal of Machine Learning Research
  9 (2008) 1871--1874.

\bibitem{Turk1991}
M.~Turk, A.~Pentland, Eigenfaces for recognition, Journal of Cognitive
  Neuroscience 3~(1) (1991) 71--86.

\bibitem{Grant2008}
M.~Grant, S.~Boyd, Graph implementations for nonsmooth convex programs, in:
  V.~Blondel, S.~Boyd, H.~Kimura (Eds.), Recent Advances in Learning and
  Control, Lecture Notes in Control and Information Sciences, Springer-Verlag
  Limited, 2008, pp. 95--110.

\bibitem{Savvides2006}
M.~Savvides, R.~Abiantun, J.~Heo, S.~Park, C.~Xie, B.~Vijayakumar, Partial
  holistic face recognition on frgc-ii data using support vector machine, in:
  Proc. of Computer Vision and Pattern Recognition Workshop, 2006, pp. 48--53.

\bibitem{Sinha2006-Recognition}
P.~Sinha, B.~Balas, Y.~Ostrovsky, R.~Russell, Face recognition by humans:
  Nineteen results all computer vision researchers should know about,
  Proceedings of the IEEE 94~(11) (2006) 1948--1962.

\bibitem{Qiao2010Sparsity}
L.~S. Qiao, S.~C. Chen, X.~Y. Tan, Sparsity preserving projections with
  applications to face recognition, Pattern Recognition 43~(1) (2010) 331--341.

\end{thebibliography}

\end{document}